\documentclass[10pt,journal,compsoc]{IEEEtran}
\usepackage{etex}
\reserveinserts{36}

\usepackage{graphicx}
\usepackage{subcaption}
\usepackage{newclude}
\usepackage{booktabs}

\usepackage{tikz,amsmath,amssymb,bm,color}
\usetikzlibrary{calc,trees,positioning,arrows,shapes,chains,shapes.geometric,decorations.pathreplacing,decorations.pathmorphing,shapes,matrix,shapes.symbols}
\usetikzlibrary{external}
\usetikzlibrary{plotmarks}
\usepackage{filecontents}
\usepackage{pgfplots,pgfplotstable}
\usepackage{bm}
\usetikzlibrary{intersections}

\usepackage{url}
\usepackage{soul}		
\usepackage[normalem]{ulem}

\usepackage[colorinlistoftodos]{todonotes}
\makeatletter
\renewcommand{\todo}[2][]{\tikzexternaldisable\@todo[#1]{#2}\tikzexternalenable}
\makeatother

\usepackage[finalnew]{trackchanges}
\addeditor{CL} 
\addeditor{RP} 
\addeditor{JS} 

\ifCLASSOPTIONcompsoc
\usepackage[nocompress]{cite}
\else
\usepackage{cite}
\fi
\DeclareMathOperator*{\argmin}{arg\,min}

\begin{document}

\title{Sparsity in Dynamics of Spontaneous \\Subtle Emotions: Analysis \& Application}

\author{Anh~Cat~Le~Ngo,~\IEEEmembership{Member,~IEEE,}
        John~See,~\IEEEmembership{Member,~IEEE,}
        Raphael C.-W. Phan,~\IEEEmembership{Member,~IEEE}%
\thanks{Anh Cat Le Ngo and Raphael C.-W. Phan are with the Faculty of Engineering, Multimedia University, Malaysia.}

\thanks{John See is with the Faculty of Computing \& Informatics, Multimedia University, Malaysia.}}


\IEEEtitleabstractindextext{%
\begin{abstract}
Subtle emotions are present in diverse real-life situations: in hostile environments, enemies and/or spies maliciously conceal their emotions as part of their deception; in life-threatening situations, victims under duress have no choice but to withhold their real feelings; in the medical scene, patients with psychological conditions such as depression could either be intentionally or subconsciously suppressing their anguish from loved ones. 
Under such circumstances, it is often crucial that these subtle emotions are recognized before it is too late. 
These spontaneous subtle emotions are typically expressed through micro-expressions, which are tiny, sudden and short-lived dynamics of facial muscles; thus, such micro-expressions pose a great challenge for visual recognition. 
The abrupt but significant dynamics for the recognition task are temporally sparse while the rest, i.e. irrelevant dynamics, are temporally redundant. \change[CL]{In this work, we decompose the dynamics into spatial and temporal dynamic modes at several specific spectral frequencies; then, enforce sparsity constraints on the analysis to learn temporal and spectral structures of sparsity in the dynamics. These temporal and spectral sparsity-promoting analyses enable to reveal significant and eliminate irrelevant facial dynamics of micro-expressions which ultimately would ease the challenge in the visual recognition of spontaneous subtle emotions.}{In this work, we analyze and enforce sparsity constraints to learn significant temporal and spectral structures while eliminating irrelevant facial dynamics of micro-expressions, which would ease the challenge in the visual recognition of spontaneous subtle emotions.} The hypothesis is confirmed through experimental results of automatic spontaneous subtle emotion recognition with several sparsity levels on CASME II and SMIC, the \remove[CL]{only} two \add[CL]{well-established and} publicly available spontaneous subtle emotion databases. The overall performances of the automatic subtle emotion recognition are boosted when only significant dynamics of the original sequences \add[JS]{are preserved}.
\end{abstract}

\begin{IEEEkeywords}
	Spontaneous subtle emotions, emotion suppression, data sparsity, dynamic mode decomposition, micro-expression recognition.
\end{IEEEkeywords}}
	
\maketitle

\IEEEdisplaynontitleabstractindextext

%
\IEEEpeerreviewmaketitle





\IEEEraisesectionheading{\section{Introduction}\label{sec:itr}}

\IEEEPARstart{S}{ubtle} emotions are leaked through micro-expressions of human faces which are mostly spontaneous and unintentional but highly informative. As these expressions are mostly ignited in the sub-conscious part of the brain, it reveals the true emotional state of the subjects. As toddlers grow up to become adults and society evolves through the ages, human beings become more complex, unpredictable and difficult to read. People learn to control and suppress their voluntary expressions. Equally, the task of spotting and recognizing these spontaneous micro-expressions has become more difficult and challenging. Even trained human experts only recognize 47\% of samples from 5-class subtle emotion recognition tasks correctly, while accuracy of novices is much lower at approximately 32\% \cite{frank2009see}. The challenges and limitations of human capability in performing this recognition task motivates researchers in various interdisciplinary fields such as computer vision, neuro-computing and psychology to design feasible schemes and methods for automatic recognition of spontaneous subtle emotions \cite{park2009subtle,wu2011machine,pfister2011recognising,yan2014micro,lengo2014imbalance}. The range of potential applications is plentiful, in particular--- lie and deception detection \cite{ekman2009lie,warren2009detecting}, remediation of depression and schizophrenia \cite{surguladze2004recognition,russell2006pilot} and speech-making influence \cite{stewart2009presidential}.
 
These subtle emotions, also known as \emph{micro-expressions}, typically last for approximately 1/25 to 1/15 of a second \cite{ekmanMicro}, with a generally acceptable upper limit of 1/2 a second \cite{yan2013fast}. Their intensities are small, indiscriminate and localized at times. This makes recognizing subtle emotion by watching normal frame-rate video recordings through the naked eye difficult as high-speed cameras are able to capture more data from such subtle changes. In \cite{ekman2009lie}, Ekman suggested that human evaluators can first view the video frame-by-frame and then repeat the process with increasing speed to aid the identification of type of micro-expression.

However, \emph{more} generated data does not necessarily translate to \emph{more} information as there are possibilities that redundant information and/or noise may have been unwittingly captured as well. Spontaneous expressions (by way of being naturalistic affective expressions \cite{meng2014affective}) are by nature, less stereotypical and full-fledged than typical voluntary expressions. They tend to change more slowly than acted out expressions. As such, consecutively recorded instants of expressions may after all, represent the same information. In other words, the dynamic activities of subtle emotions are possibly sparse in video samples recorded at high frame rates (measured by \emph{fps} or frames-per-second). This hypothesis is supported by analyzing a landmark work by Pfister et al. \cite{pfister2011recognising} which introduced the tasks of detecting and recognizing facial micro-expressions using a newly-created SMIC corpus. Though video samples in the SMIC corpus were captured at 100 \emph{fps} with an average sample frame length of 29, they were all normalized to 10 frames to achieve the best performance in both detection and recognition tasks. Nevertheless, this apparent sparsity is not only dealt with in a high-speed corpus like SMIC. Further experiments by Pfister et al. \cite{pfister2011recognising} on a much slower YorkDDT corpus (at 25 \emph{fps}) \cite{warren2009detecting}, also yielded a significant boost in detection and recognition rate when the sample lengths were normalized to just 10 frames long. Generally, this points towards the importance of dealing with the sparsity of dynamics found in micro-expression databases, and that good utilization of this property can in turn produce positive results in detection and recognition tasks.  

Temporal interpolation model (TIM) of image sequences was first introduced by Pfister et al. \cite{pfister2011recognising} for facial expression data, in order to synthesize sufficient frames for extracting spatiotemporal local texture descriptors. Interestingly, Pfister et al. pointed out the fact that their best performances for detection and recognition on both datasets (YorkDDT and SMIC) were achieved by temporally downsampling the video sequences to a fixed length of 10 frames, with an obvious deterioration of performance due to redundant data when more frames were considered. However, they stopped short of analyzing this redundancy further. Moreover, TIM neither offers any rubrik as to how and why a certain number of frames should be sampled from the original sequences, nor does it help to visualize dynamics of these subtle movements. Prior to that, Zhou et al. \cite{zhou2011towards} also proposed a similar graph embedding method for interpolating frames for synthesizing a talking mouth.
\todo[size=\footnotesize]{Raph asks: add elaboration on what's the limitation.}

In this paper, we aim to analyze and visualize the sparsity and redundancy of dynamics in micro-expressions to improve recognition of spontaneous subtle emotions. 
For extraction and visualization of the dynamics, we base our approach on
the concept of Dynamic Mode Decomposition (DMD) \cite{schmid2010dynamic}, a popular technique in fluid dynamic analysis which was recently adopted for video processing research \cite{grosek2014dynamic}. DMD originally aims to model changes between consecutive discrete video frames of continuous fluid flow; therefore, we show that this technique can be applied to analyze micro-expressions which are dynamically changing slowly over time. Moreover, DMD is capable of analyzing signals in both temporal and spectral domains simultaneously, by representing data as a linear combination of modes weighted by their amplitudes at a particular instance in time.
In other words, it is able to localize the signals at a specific time and frequency (mode), which is not possible in most signal analysis schemes like Fourier Transform or Wavelet due to Heisenberg uncertainty principle \cite{mallat1999wavelet}. In this work, we utilize the recently proposed sparsity-promoting DMD (DMDSP) \cite{jovanovic2014sparsity} technique which enforces sparsity constraint to classical DMD analysis. It is capable of reducing redundant data while still preserving the most significant and discriminating dynamics of each video sequence. Through a series of experiments on subtle emotion recognition, we show that the elimination of redundant data by enforcing sparsity constraints can boost recognition performance. Furthermore, we also demonstrate the effectiveness of DMDSP in capturing the essential dynamic modes in micro-expressions as compared to temporal interpolation model (TIM). In brief, the contributions of this paper are summarized as follows:
\todo[size=\footnotesize]{Revise after going through the whole paper}
\begin{enumerate}
\item Visualization of micro-expressions' dynamics in temporal and spectral domains with DMD
\item Sparsity analysis of micro-expressions' dynamics in temporal and spectral domains with DMDSP
\item Theoretical comparisons between down-sampling of TIM and sparsity enforcing of DMDSP in removals of redundant data
\item Comparison in performances of subtle emotion recognition systems with and without elimination of dynamic redundancy.
\item Description of the best parameters and proposed methods for redundancy removals in this paper.
\end{enumerate}

\section{Introduction}
\label{sec-bkg}
\change[CL]{In our current interconnected world where we interact on social media and perform our daily tasks increasingly via our mobile devices, the user experience trend is now towards human-centric designs, where computers and devices are to perceive and understand humans rather than vice versa.}{In our current era of social networks facilitated primarily by the widespread accessibility of internet-ready on-person mobile devices, smart human-centric systems are increasingly expected to perceive and understand humans rather than vice versa.} Instead of simply executing users' commands, computers need to understand the \change[CL]{many forms}{multi-modality} of human-like communications. This includes recognition of facial expressions, the non-verbal form of human communications. The shift in paradigm toward human-friendly computing has initiated the field of Affective Computing \cite{picard2010affective}. This section briefly surveys recent methods and advances in an emerging subfield of Affective Computing: automatic recognition of subtle emotions through facial micro-expressions. Though automatic recognition of spontaneous subtle emotions is a new challenging task, recognition systems for normal expressions have been a research subject for \change[CL]{more than 15 years}{nearly two decades}. Shan et al. \cite{shan2009facial} \change[CL]{highlighted}{summarized achievements of this research field} in a popular framework of automatic facial expression and emotion recognition systems\change[CL]{and then}{. Furthermore, the authors} focused on \add[CL]{analyzing} two core components: facial representations e.g. LBPTOP, and classifiers e.g. SVM and AdaBoost. For normal expressions, this framework successfully recognizes with\remove[CL]{recognition rates of} above 90\% \add[CL]{accuracy} \cite{shan2009facial}. However, the same framework falls short of this impressive \change[CL]{accuracy}{recognition rate} when dealing with micro-expressions \cite{yan2014casme,lengo2014imbalance}.

\begin{figure}
		\includegraphics[width=\linewidth]{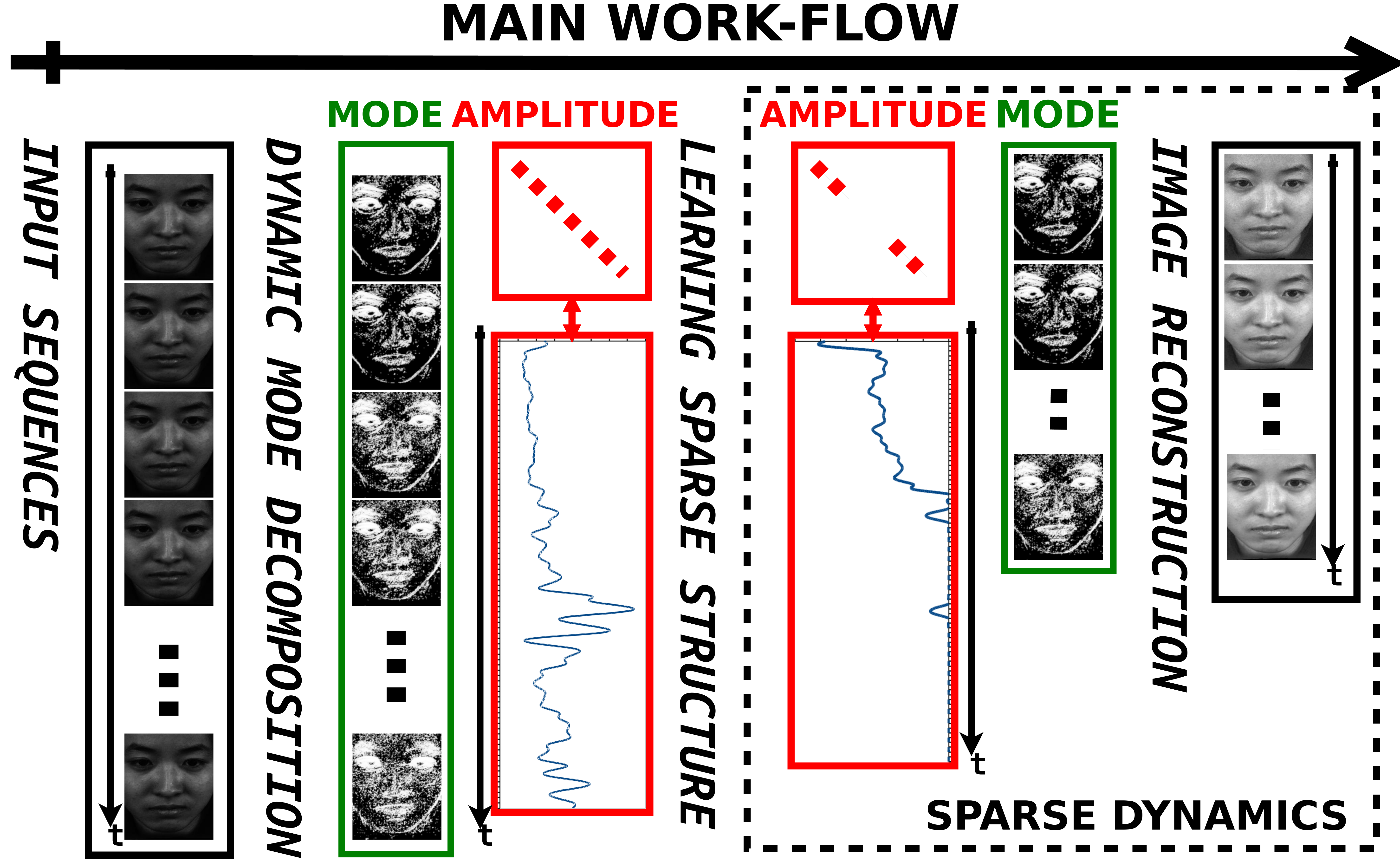}
	\caption{Visualization of steps in Sparse Promoting Dynamic Mode Decomposition (DMDSP) for removals of redundant neutral faces: (1) Input sequences are analyzed by Dynamic Mode Decomposition (DMD), (2) $L_1$ sparsity is applied to achieve the least number of modes and give the minimum \change[CL]{of losses up}{loss} during reconstruction, (3) Output sequences are reconstructed from sparse modes.}
	\label{sec-bkg:fig-1}
\end{figure}

As micro-expressions of subtle emotions are much more elusive than normal expressions due to their small intensities, short-liveness (between $\frac{1}{25}$s and $\frac{1}{15}$s \cite{ekmanMicro}) and unpredictability, their image sequences need to be pre-processed to reduce these unfavorable characteristics. In this paper, we firstly propose removal of redundant neutral faces from micro-expression sequences and keeping only sparse and significant frames, which is illustrated in Figure \ref{sec-bkg:fig-1}.
As the sparse frames significantly contribute to reconstruction of original dynamics; meanwhile, redundant frames could be omitted without much cost i.e. errors between reconstructed and original sequences. The proposed pre-processing technique aims to remove as many neutral and redundant frames as possible with minimum cost. It not only produces more visually distinguishable but also allows extraction of more discriminant features. As a result, it improves the accuracy rate of automatic subtle emotion recognition. Secondly, we carry out temporal and spectral analysis of subtle emotion sequences so as to clarify rationales behind our approaches as well as select suitable experimental parameters. Finally, we compare performances of our proposed solution with those of the state-of-the-art methods on recognition of spontaneous micro-expressions. 


Section \ref{sec-bkg:subsec-dbs} describes two publicly available spontaneous subtle emotions databases: CASME II \cite{yan2014casme} and SMIC \cite{pfister2011recognising}, which are utilized as input data for our experiments. As system performance greatly depends on characteristics of these databases, understanding \add[CL]{their} pros and cons \change[CL]{of the database and their}{as well as} samples provides knowledge and clues for designing \change[CL]{appropriate}{optimal} automatic recognition systems. In Section \ref{sec-bkg:subsec-rlw}, related works in recent literatures are reviewed and categorized with respect to their main contributions in preprocessing, feature extraction or classification stages. Section $\ref{sec-dpm}$ elaborates details of dynamic preprocessing techniques: Temporal Interpolation Method (TIM) \cite{pfister2011recognising}, Dynamic Mode Decomposition (DMD) \cite{schmid2010dynamic} and Sparsity-promoting Dynamic Mode Decomposition (DMDSP) \cite{jovanovic2014sparsity}. Meanwhile, Section \ref{sec-dan} utilizes DMD magnitudes to analyze responses of temporally dense (TIM) and sparse (DMDSP) sampling approaches in the temporal and spectral domains. Section \ref{sec-eds} describes with what parameters and how the proposed method \change[CL]{is}{are} evaluated as well as compared to both baselines \cite{yan2014casme}\cite{li2013spontaneous} and other recent state-of-the-art methods \cite{huang2015facial},\cite{oh2015monogenic},\cite{liong2014subtle}. Finally, conclusions are drawn in Section \ref{sec-con}.


\subsection{Spontaneous Subtle Emotion Databases}
\label{sec-bkg:subsec-dbs}
Various facial expression databases had been proposed in the literature. However, little attention has been paid to \emph{spontaneous} micro-expression databases, partly due to difficulties faced in proper elicitation of samples and labeling by experts. Hence, the lack of well-established databases for spontaneous micro-expression researches poses a challenge towards the design of automatic detection and recognition systems. 
There is a need to emphasize on the actual meaning of the term "spontaneous micro-expressions", in contrast to what had been previously regarded generally as "micro-expressions". Ekman's \cite{ekmanMicro} and Yan et al.~\cite{yan2014casme} suggest that micro-expressions should be considered involuntary and difficult to disguise. Previous micro-expression databases such as the USF-HD~\cite{shreve2011macro} and Polikovsky's~\cite{polikovsky2009facial} databases contained micro-expressions that were actually posed or acted out instead of naturally spontaneous ones. Moreover, the occurrence duration of their micro-expressions were longer ($\frac{2}{3}$s) than Paul Ekman's definition ($\frac{1}{3}$s) \cite{ekmanMicro}. Another database is the YorkDDT~\cite{warren2009detecting}, which includes the micro-expression in a spontaneous manner\change[CL]{ but }{. However,} there were other irrelevant facial and head movements therein, thus complicating the recognition process. Besides, \change[CL]{it}{YorkDDT} contained only 18 micro-expressions which are insufficient for proper experimentation and analysis.  

However, there are two recent and comprehensive databases that meet the requirements of spontaneous micro-expressions, namely the SMIC\footnote{\url{http://www.cse.oulu.fi/SMICDatabase}}~\cite{pfister2011recognising} and CASME II\footnote{\url{http://fu.psych.ac.cn/CASME/casme2-en.php}}~\cite{wang2014micro} (an improved extended version of the original CASME by the same researchers). CASME II has 247 video samples from 26 subjects and SMIC has 164 samples from 16 subjects. They are both publicly available and contain sufficiently large number of video samples which are conducive for a micro-expression recognition research. \change[CL]{Since micro-expression recognition research is at an early stage, b}{B}oth databases were recorded in a constrained laboratory condition \change[CL]{with the labeling done}{annotated} by two trained \change[CL]{annotators (or coders)}{coders} and also the participants' self-reports. Non-emotional facial movements were also eliminated from the final selected sequence. Samples from both SMIC and CASME II were acquired from relatively high frame rates (100 fps and 200 fps respectively) to better locate the occurrence of micro-expressions.

\subsection{Related Work}
\label{sec-bkg:subsec-rlw}
\change[CL]{This section discusses the}{Discussion about} recent updates in the three main stages of the subtle emotion recognition system: pre-processing, facial representation and classification, \add[CL]{is presented} in the following subsections \ref{sec-bkg:subsec-rlw:subsubsecf-sep}, \ref{sec-bkg:subsec-rlw:subsubsecf-sef}, and \ref{sec-bkg:subsec-rlw:subsubsecf-sec} respectively.
\subsubsection{Subtle Emotion Preprocessing}
\label{sec-bkg:subsec-rlw:subsubsecf-sep}
The subtleness of spontaneous emotions is challenging to be recognized due to two main problems: small dynamics of facial muscles, and involuntary and unexpected expressions. Therefore, video samples need pre-processing steps to better visualize \remove[CL]{the} changes in subtle emotions and subsequently
extract more distinctive features. As motions of facial muscles in micro-expressions are too small, Le Ngo et al. \cite{lengo2016eulerian} \change[JS]{shows}{showed} that motion magnification techniques \cite{wu2012eulerian}\cite{wadhwa2013phase}\cite{liu2005motion} improve \add[JS]{the} recognizability of these expressions. \change[JS]{The}{These} techniques \change[JS]{would}{are able to} increase the emotional intensity of micro-expressions, making them more visible like normal expressions. Moreover, this magnification effect can be achieved by fast Eulerian Motion Magnification techniques \cite{wu2012eulerian}\cite{wadhwa2013phase} instead of the Lagrangian approach \cite{liu2005motion} which often requires motion estimations, and other \change[CL]{heavy computations}{heavily computational processes}. However, magnification of micro-expression is \change[JS]{out of}{outside the} scope of this work as this paper mainly focuses on the sparsity of these expressions.

While motion magnification deals with small displacements between frames of micro-expression video samples, temporal interpolation method (TIM) \cite{pfister2011recognising} tackles the unexpectedness of micro-expressions. Bursts of spontaneous subtle emotions are difficult to be detected accurately; therefore, video samples are often cut from a long recording of a subject's expressions. While on-set and off-set points are identified by trained experts in subtle emotions to indicate the starting and ending of a micro-expression sequence, these points are hardly accurate as well. Therefore, these video samples may include frames of almost neutral faces among frames of micro-expressions. \change[JS]{Moreover,}{Since} micro-expressions only last for a very short duration,\remove[JS]{thus,} neutral faces may dominate a large portion of some sequences. For a sample with many redundant neutral faces, TIM is able to interpolate at arbitrary points along \change[CL]{the}{a} temporal axis according to an embedded graph in a manifold, \add[CL]{which is in turn} learned from \change[CL]{images of the sample}{video frames}. TIM was initially aimed at synthesizing more frames, as video samples recorded at standard 25 fps were too short for subtle emotion recognition. However, the same technique could also be used to interpolate less frames or to remove redundant neutral faces, as video samples \add[CL]{were} recorded at 100 fps or 200 fps are too long. As TIM assumed that \add[RP]{facial expressions change} across consecutive frames, \add[JS]{and are }sampled along a simple graph on a manifold, it \change[CL]{was}{is} difficult to control how significant or redundant the dynamics are after the interpolation. Therefore, 
the positive effectiveness of TIM on the performance of the recognition system cannot be guaranteed. 

A technique capable of extracting coherent structures and significant dynamics at a single temporal frequency\add[CL]{,} is Dynamic Mode Decomposition (DMD) \cite{schmid2010dynamic}. \change[JS]{The technique}{DMD} \change[CL]{was}{is} a popular \add[JS]{technique} in fluid dynamics imagery, and \change[JS]{later}{it was recently} applied to foreground motion segmentation in video processing \cite{grosek2014dynamic}. A more recent variant of it, Sparsity-Promoting Dynamic Mode Decomposition (DMDSP) \cite{jovanovic2014sparsity} puts the decomposition under sparse constraints such that the least number of DMD modes are utilized for construction of original sequences. The notion of analyzing a sequence of images into more meaningful temporal structures is potentially useful; an idea which we aim to exploit in this paper.

\subsubsection{Subtle Emotion Features}
\label{sec-bkg:subsec-rlw:subsubsecf-sef}
Systems for automatic recognition or detection of micro-expressions inherited many components from those for normal expressions (or so-called macro-expressions), including \remove[CL]{the} use of features. As Local Binary Pattern with Three Orthogonal Planes (LBPTOP) \cite{zhao2007dynamic} is a common and effective feature for representing normal facial expressions \cite{shan2009facial}, it has also been utilized in several works relating to micro-expressions \cite{yan2014casme,pfister2011recognising,lengo2014imbalance}. LBPTOP is a spatiotemporal feature, encoding textural features along three orthogonal physical planes $XY$, $YT$, and $XT$ into binary sequences, where $X,Y$ are two axes of the spatial domain and $T$ is the temporal axis. The binary sequences are later summarized and concatenated in a histogram, which forms the LBPTOP feature. Local Spatiotemporal Directional Features (LSDF) was recently proposed by Wang et al. \cite{wang2014micro} for automatic recognition of subtle emotions. Instead of using the center pixel of the neighborhood for thresholding, LSDF encodes each plane along the horizontal and vertical directions. Their experiments demonstrated that LSDF was comparable to LBPTOP in most cases, if not better under certain conditions.

Besides statistical spatiotemporal textural features e.g. LBPTOP, there are other potential feature extraction and representations for micro-expressions based on optical flow, multi-scale wavelet analysis, etc. For instance, Liong et al. \cite{liong2014subtle} utilized optical strain, a derivative of optical flow, for the recognition task; building on similar concepts used for expression spotting by Shreve et al. \cite{shreve2009towards,shreve2011macro}. Furthermore, Liu et al. \cite{liu2015main} encoded statistical information of the main directional optical flows in regions of interests (ROI),  
\remove[JS]{In this method \protect\cite{liu2015main},} 
which are manually defined with respect to facial landmarks. A recent work by Oh et al. \cite{oh2015monogenic} introduced a multi-scale Riesz wavelet representation for micro-expressions that capture\change[CL]{s}{d} the monogenic signal components, i.e. magnitude, phase and orientation. Their method reported an improvement over the LBPTOP and spatiotemporal local monogenic binary pattern (STLMBP) \cite{huang2014improved}.
Furthermore, Oh et al. \cite{oh2016intrinsic} \change[JS]{experimentally proves}{proved experimentally} that intrinsic 2-D features are better than its 1-D counterpart for encoding facial micro-expressions. Recently, Huang et al. \cite{huang2015facial} have proposed a new spatio-temporal feature based on integral projection of difference images along horizontal and vertical directions, which achieved \add[JS]{a good recognition performance relative to current state-of-the-art methods.}
Despite different approaches, all features aim to extract error-prone dynamics and tiny motions of micro-expressions, which are discriminative features for recognizing subtle emotions.

\subsubsection{Subtle Emotion Classifier}
\label{sec-bkg:subsec-rlw:subsubsecf-sec}
Besides features, \remove[CL]{the} choices of classifiers for micro-expression detection and recognition are inherited from approaches for \add[JS]{normal} macro-expressions. Shan et al. \cite{shan2009facial} highlighted two popular classifiers for emotion recognition systems: Support Vector Machine \add[CL]{SVM} and AdaBoost, \add[JS]{both of} which have also been utilized for recognizing subtle emotions \cite{yan2014casme,lengo2014imbalance}. However, there is a \change[JS]{distinctive}{distinct} difference between the distribution of samples in macro- and micro-expression databases, which directly affects performance and choices of classifiers. Since video samples of spontaneous macro-expressions are widely available with a large number of samples, it is easy to get a balance between classes of macro-expressions. It is much more difficult to acquire a balanced number of spontaneous micro-expression \add[JS]{samples} due to various reasons, viz. its natural characteristic (small intensities, unexpected dynamics) and the difficulty in \change[JS]{collection}{eliciting certain emotions}. Therefore, imbalance of samples across classes is unavoidable. To tackle this imbalance, Le Ngo et al. \cite{lengo2014imbalance} \change[JS]{adapted}{introduced} an Adaboost-based person-specific classifier, and advocated the use of F1-score, precision and recall metrics in place of the conventional recognition accuracy. Their experimental results showed an improvement by a small margin when compared to standard classifiers like SVM and AdaBoost. 

\section{Dynamically Preprocessing Methods}
\label{sec-dpm}
\begin{figure}[h]
	\begin{subfigure}{0.32\columnwidth}
		\includegraphics[width=\textwidth]{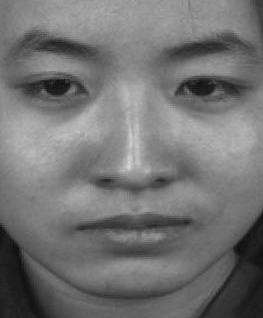}
		\caption{Subtle Emotion}
	\end{subfigure}
	\begin{subfigure}{0.32\columnwidth}
		\includegraphics[width=\textwidth]{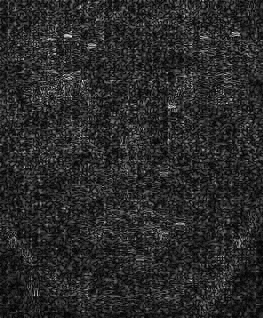}
		\caption{DMD Mode}
	\end{subfigure}
	\begin{subfigure}{0.32\columnwidth}
		\includegraphics[width=\textwidth]{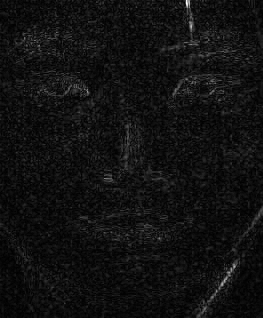}
		\caption{DMDSP Mode}
	\end{subfigure}
	\caption{Visual comparison between a noisy and redundant DMD spatial mode in (b) and a clear and significant DMDSP spatial mode in (c) from a subtle expression in (a) }
	\label{sec-dpm:subsec-dmdsp:fig-2}
\end{figure}
Most subtle emotions happen very briefly in a short period of time; therefore only high-speed recording is able to capture their full dynamics. Moreover, these expressions usually appear unexpectedly and their (beginning) on-set and (ending) off-set points are difficult to be identified exactly even by trained experts. As a result, more unnecessary frames, containing no emotional expressions, are accidentally acquired for a micro-expression sample. The redundancy is inevitable, as shown in the Figure \ref{sec-dpm:subsec-dmdsp:fig-2} visualization of redundant and significant spatial modes of a subtle expression. Due to the inseparability of identities and emotions, unnecessary neutral faces only confuse classifiers of subtle expressions and dampen performances in the \add[CL]{latter} recognition task. Hence, removal of these undesired frames is crucial. 

\begin{figure}[t]
	\begin{subfigure}{\columnwidth}
		\IfFileExists{sample_signal.tex}
		{\input{sample_signal.tex}}
		{\includegraphics[width=\columnwidth]{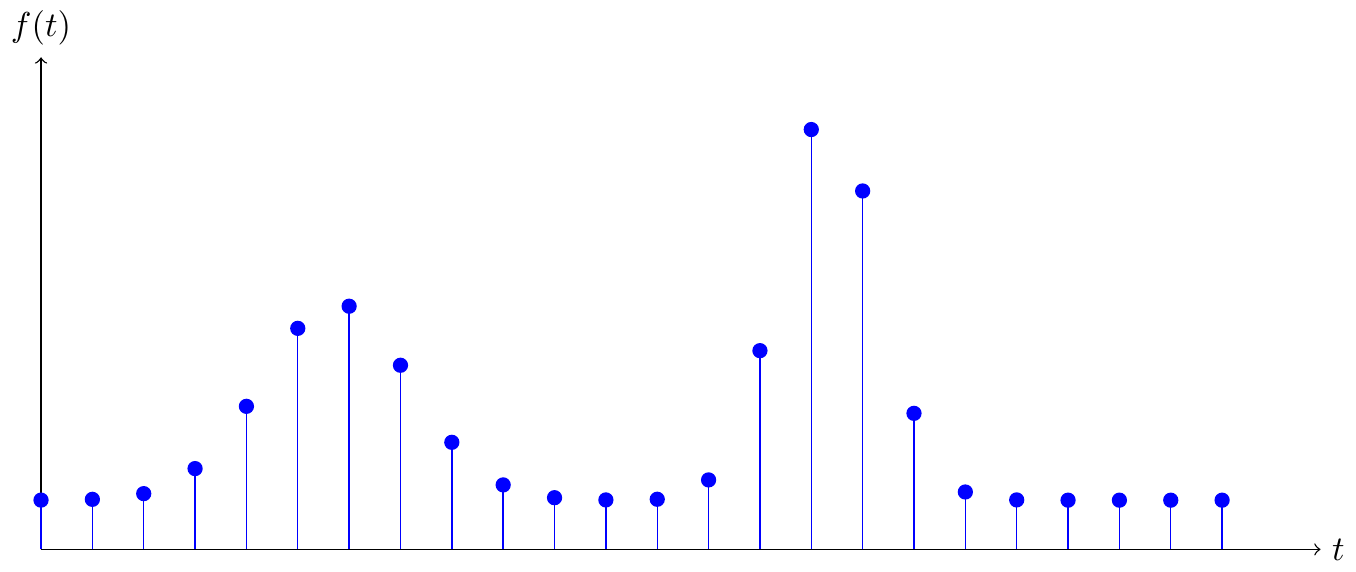}}
		\captionof{figure}{Sample Signal}
		\label{sec-dpm:fig-1}
	\end{subfigure}
	\begin{subfigure}{\columnwidth}
		\IfFileExists{tim_signal.tex}
		{\input{tim_signal.tex}}
		{\includegraphics[width=\columnwidth]{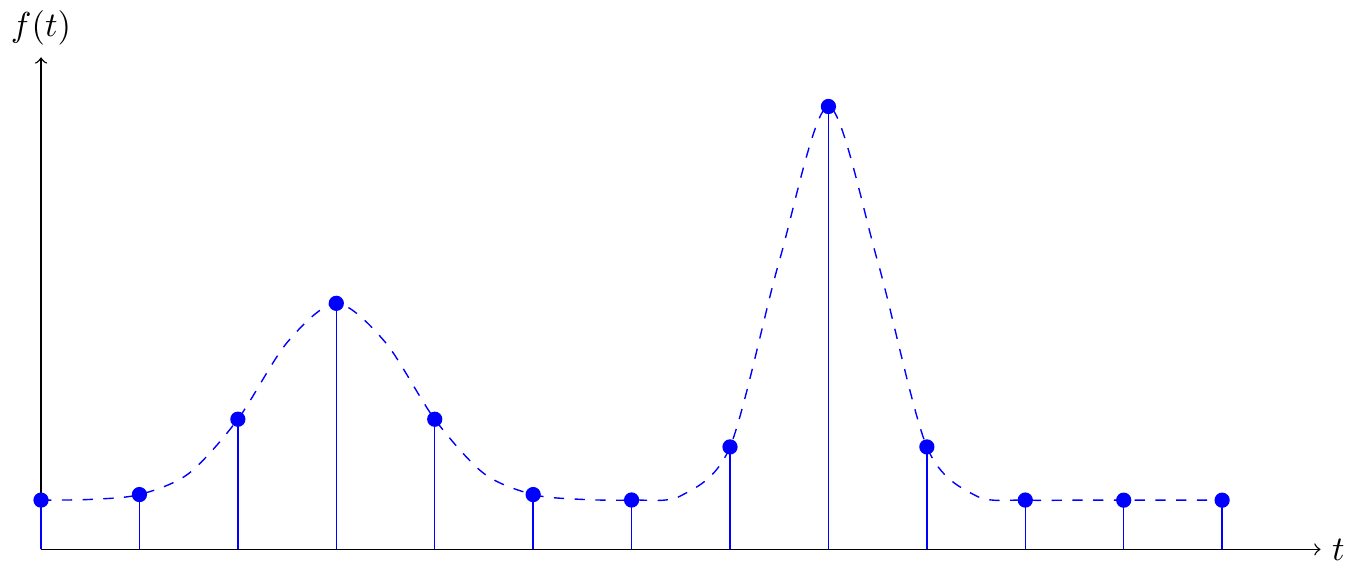}}
		\captionof{figure}{TIM Pre-processed Signal}
		\label{sec-dpm:subsec-tim:fig-1}
	\end{subfigure}
	\begin{subfigure}{\columnwidth}
		\IfFileExists{dmdsp_signal.tex}
		{\input{dmdsp_signal.tex}}
		{\includegraphics[width=\columnwidth]{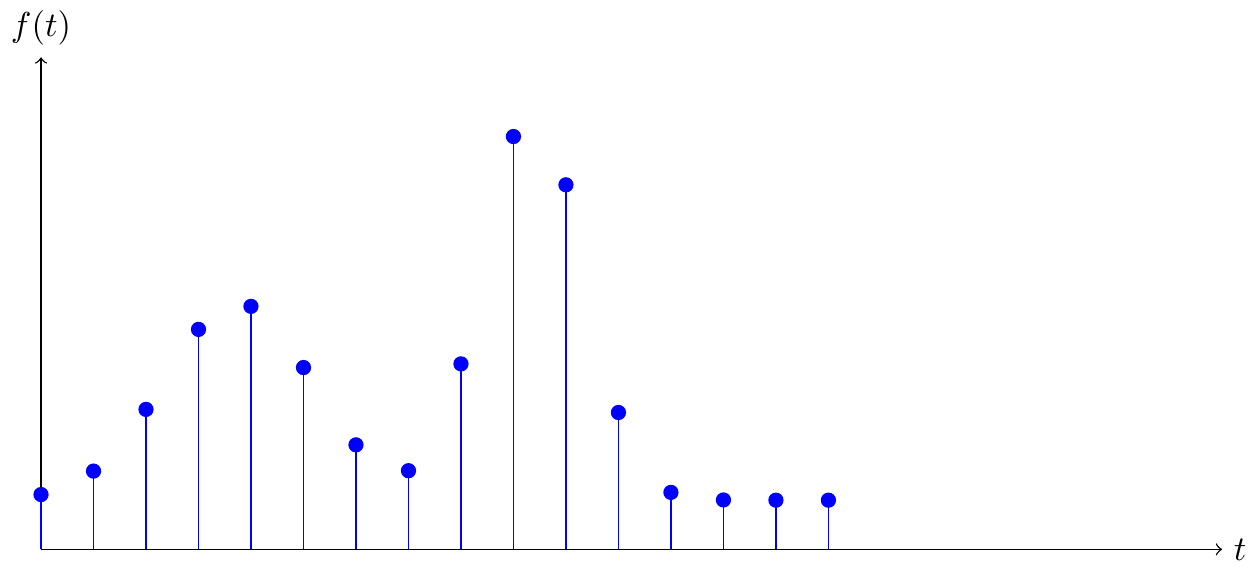}}
		\captionof{figure}{DMDSP Pre-processed Signal}
		\label{sec-dpm:subsec-dmdsp:fig-1}
	\end{subfigure}
	\caption{Demonstration of how a sample signal (a) is pre-processed by TIM (b) and DMDSP (c) where the vertical axis, $f(t)$, represents the magnitude of dynamics at the horizontal axis $t$.}
\end{figure}

Lets consider a discrete signal, in Figure \ref{sec-dpm:fig-1}, which represents dynamic magnitudes $f(t)$ of a spontaneous subtle emotion at time $t$. This toy example deliberately demonstrates the redundancy assumption of the dynamics $f(t)$ as only two short parts of the sample signal have significant magnitudes while the majority of these discrete signals have relatively small magnitudes. In other words, significant dynamics are sparse and insignificant ones are redundant for reconstruction of facial dynamics as most signal energy is concentrated into these two local peaks. We hypothesize that micro-expressions would become more descriptive and discriminative if \change[CL]{samples only keep}{only} significant dynamics \add[CL]{are kept} and\remove[CL] {eliminate} redundant ones \add[CL]{are eliminated}. In the following Sub-sections \ref{sec-dpm:subsec-tim} and \ref{sec-dpm:subsec-dmdsp}, we discuss two approaches to deal with this redundancy -- Uniform Sampling and Sparse Sampling approaches; the latter being our proposed scheme.

\subsection{Uniform Sampling Approach: Temporal Interpolation Method (TIM)}
\label{sec-dpm:subsec-tim}

In this approach, micro-expressions are assumed to happen continuously and 
sampled along a curve on a low-dimensional manifold. If the curve and its manifold are successfully parameterized, a specific number of frames can be synthesized or interpolated from the original video frames. Figure \ref{sec-dpm:subsec-tim:fig-1} demonstrates how this so-called \emph{uniform selection} approach fits the original discrete signals in a curve (dashed line) and synthesizes samples at arbitrary but equispaced points along that curve (circular markers). The synthesized discrete signal can either interpolate towards a reduced number of samples (as seen in Figure \ref{sec-dpm:subsec-tim:fig-1}), or extrapolate to more samples if necessary.

One example of an approach that performs the described uniform selection is the Temporal Interpolation Method (TIM), used by Pfister et al. \cite{pfister2011recognising} and first suggested by Zhou et al. \cite{zhou2011towards}. While Zhou et al.\remove[CL]{aimed to} extrapolated frames for practical lip-reading\change[CL]{and}{,} Pfister et al. \change[CL]{extrapolated frames}{aimed} for recognizing subtle emotions from normal frame-rate recorded sequences. Both applications aimed to synthesize proper frame-lengths for \remove[CL]{. This enables} stable spatio-temporal feature extraction\remove[CL]{under such scarcity}. TIM was utilized by Pfister et al. in the opposite way (interpolation) instead of extrapolation. As spontaneous subtle emotions are recorded with high frame-rate, TIM interpolates fewer frames than original frame-lengths to remove redundancy in the dynamics of micro-expressions. Interpolated frames are uniformly sampled at \remove[CL]{arbitrary but} equispaced positions of a graph embedded on a manifold. These frames are assumed to be represented by vertices on a path graph $P_n$, where $n$ is the frame number. \remove[CL]{of frames or vertices.} The relationship between adjacent frames \remove[CL]{are}{is} modeled by the adjacency matrix $\mathbf{W} \in \{0,1\}^{n \times n}$ with $W_{i,j}=1$ if they are two consecutive frames, $\|i-j\|=1$, and 0 otherwise. The graph $P_n$ lies on a manifold when the total length of edges between connected vertices is minimized according to the following equation,

(\ref{sec-dpm:subsec-tim:eqn-1}):
\begin{align}
\sum_{i,j} (y_i - y_j)^2 W_{i,j}, \quad i,j=1,2,\ldots,n
\label{sec-dpm:subsec-tim:eqn-1}
\end{align}
where $\mathbf{y}=(y_1,y_2,\ldots,y_n)$ is an eigenvector of the Laplacian graph of $P_n$. All points on an eigenvector $\mathbf{y_k}$ are assumed to stay on a sinusoidal graph, formulated by, 
\begin{align}
f_k^n(t) = \sin(\pi k t + \pi (n-k) / (2n)), t \in [1/n,1]
\label{sec-dpm:subsec-tim:eqn-2}
\end{align}
For $n$ vertices, the Laplacian graph has $n-1$ eigenvectors $\{\mathbf{y_1},\mathbf{y_2},\ldots,\mathbf{y_{n-1}}\}$. In general, the manifold can be characterized with a collection $F^n(t)$ of $n-1$ sinusoidal curves; moreover, frames at arbitrary positions are able to be interpolated as well. As each vertex in the graph $P_n$ corresponds to a frame of an image sequence, the specific $F^n$ of that sequence can be parametrized by mapping each frame to each point, defined by $F^n(1/n),F^n(2/n),\ldots,F^n(1)$. Moreover, the parameterization requires linear extension of graph embedding \cite{yan2007graph} in which a transformation $\mathbf{w}$ is learned to minimize the Equation \ref{sec-dpm:subsec-tim:eqn-1} as follows:
\begin{align}
\argmin_{w} \sum_{i,j} (\mathbf{w^T x_i - w^T x_j})^2 W_{i,j}
\label{sec-dpm:subsec-eqn3}
\end{align}
where $x_i = \varepsilon_i - \bar{\varepsilon}$ is a mean-removed vector and $\varepsilon_i$ is a vectorized image. He et al. \cite{he2005neighborhood} reformulated the minimization as an eigenvalue problem and solved it by singular value decomposition. More details about this minimization can be found in \cite{zhou2011towards}. As mapping with frames of an image sequence defines a specific set of curves $F^n(t)$, synthesis of new frames at arbitrary temporal points is shown by Zhou et al. \cite{zhou2011towards}. The synthesized frames seem to be a temporally smoothened version of the original sequence \cite{pfister2011recognising}. 


\subsection{Sparse Sampling Approach: Sparsity-Promoting Dynamic Mode Decomposition (DMDSP)}
\label{sec-dpm:subsec-dmdsp}
The uniform selection approach, described in the previous Subsection \ref{sec-dpm:subsec-tim}, \change[CL]{tries to}{only} temporally and regularly down-sample\add[CL]{s} image sequences \change[CL]{so that the dominant}{but does not eliminate} redundant dynamics.
TIM uniformly samples along a low-dimensional dense manifold, which is assumed to be the representative space of frames in a video sequence. The assumption would unknowingly eliminate some sparse structures of the dynamics. Therefore, alternative techniques for dynamic analysis could be used instead of graph-embedding manifold.
\add[CL]{A \emph{sparse selection} approach tries to learn sparse structures of underlying dynamics and their appropriate magnitudes.} Dynamic Mode Decomposition (DMD), which was first designed for capturing the momentum of indefinite-dimensional systems such as fluid flow, projects the complex system onto a low-complex\add[CL]{ity} 
subspace spanned by dynamic modes with a few degrees of freedom. DMD does not make any rigid assumption about the existence of a manifold governing the overall dynamics, but freedom of \change[CL]{the}{an} achieved model is data-driven. As such, DMD is only suitable for analyzing temporal dynamics but not learning their sparse structures. Therefore, we use a sparsity-constrained variant of DMD, i.e. the Sparsity Promoting DMD (DMDSP), developed by Jovanovic et al. \cite{jovanovic2014sparsity} to select only significant dynamics of subtle expressions as visually illustrated in Figure \ref{sec-dpm:subsec-dmdsp:fig-2}. Further details of DMD and DMDSP are described in the following Sub-sections \ref{sec-dpm:subsec-dmdsp:subsubsec-dmd}, \ref{sec-dpm:subsec-dmdsp:subsubsec-dmdsp}.

\subsubsection{Dynamic Mode Decomposition}
\label{sec-dpm:subsec-dmdsp:subsubsec-dmd}
Dynamic Mode Decomposition (DMD) \cite{schmid2010dynamic} is designed to extract coherent structures at a single temporal frequency or dynamic mode e.g. flows in fluid dynamics \cite{schmid2010dynamic} and motions in surveillance videos \cite{grosek2014dynamic}. 
The DMD technique was first proposed and utilized in the analysis of fluid dynamics imagery. It analyzes sequences of ``snapshots'' from
a data matrix, which are regularly sampled from fluid motions across time. Lets denote the $N+1$ sequential frames as $\{\bm{\psi_0,\psi_1,\ldots,\psi_N}\}$; the previous frame i.e. $\bm{\psi_0}$ evolves to the next frame i.e. $\bm{\psi_1}$ over \add[CL]{a} regular temporal grid\remove[CL]{s} with a constant duration $\Delta t$. To model that evolution across all frames of an image sequence, two clusters of previous frames and next frames are formed as follows.
\[
\bm{\Psi_0 := [\psi_0,\psi_1,\ldots,\psi_{N-1}] \quad\quad \Psi_1 := [\psi_1,\psi_2,\ldots,\psi_{N}]}
\]
DMD assumes that dynamics between consecutive frames are governed by a linear time-invariant transformation $A$ such that $\bm{\psi_{t+1}=A\psi_t}$ for $t \in [0,\ldots,N-1]$. Hence,
\begin{align}
\bm{\Psi_1} &= \bm{[\psi_1,\psi_2,\ldots,\psi_N]} \\ \nonumber
			&= \bm{[A\psi_0,A\psi_1,\ldots,A\psi_{N-1}] = A\Psi_0}
\end{align}
For a rank-$r$ matrix of the cluster $\Psi_0$, the transformation $A$ can be further spanned on a proper orthogonal basis $U$ for an optimal representation $\bm{F \in C^{r \times r}}$ as follows.
\begin{align}
\bm{A \approx U F U^* \quad \Psi_0 = U \Sigma V^*}
\label{sec-dpm:subsec-dmdsp:eqn-1}
\end{align} 
where $\bm{U^*}$ is a complex conjugate transpose of the basis $\bm{U}$, which is obtained from an economy-size singular value decomposition (SVD) of $\bm{\Psi_0}$. The economy-size SVD is applicable for a tall matrix, $\bm{\Psi_0 \in C^{M\times N}}$, as the number of pixels $M$ of each frame $\psi_t$ is often many more than $N$, the number of frames. As DMD models the evolution of the cluster $\Psi_0$ into the cluster $\bm{\Psi_1}$ with a linear transformation $\bm{A}$, the evolution can be regarded as a time-invariant system $\bm{\Psi_1 = A\Psi_0}$. The transformation $\bm{A}$ can be found by minimizing the following Frobenius norm:
\begin{align}
\argmin_{\textbf{A}} \bm {\| \Psi_1 - A\Psi_0 \|^2_F} = \argmin_{\textbf{F}} \bm {\| \Psi_1 - U F \Sigma V^* \|}
\end{align}
With a few linear algebra calculations, the optimal solution for the above formula is achieved as:
\begin{align}
\bm{F_{dmd} = U^* \Psi_1 V \Sigma^{-1}}
\end{align}
As $\bm{F_{dmd}}$ is a rank-$r$ matrix, it has a full set of linearly independent eigenvectors $\{\bm{y_1},\ldots,\bm{y_r}\}$ and corresponding eigenvalues $\{\mu_1,\ldots,\mu_r\}$. Then, $\bm{F_{dmd}}$ can be expressed in diagonal form,
\begin{align}
	\bm{F_{dmd} = Y D_{\mu} Z^* =}
    \left[
    \begin{matrix} 
    	\bm{y_1} & \ldots & \bm{y_r}
    \end{matrix}
    \right]  
    \left[
    \begin{matrix}
    	\mu_1 & 	& \\
        	  & \ddots & \\
              & 	& \mu_r
    \end{matrix}
    \right]    
    \left[
    \begin{matrix}
    	\bm{z_1^*} \\
        \vdots \\
        \bm{z_r^*}	
    \end{matrix}
    \right]
    \label{sec-dpm:subsec-dmdsp:subsubsec-dmd:eqn-2}
\end{align}
where $\{\bm{z_1},\ldots,\bm{z_r}\}$ are eigenvectors of $\bm{F_{dmd}^*}$. These eigenvectors are bi-orthogonal to the $\{\bm{y_1},\ldots,\bm{y_r}\}$, which means $\bm{Z^*Y = I}$ and $\bm{F_{dmd}^{t} = Y D_{\mu}^{t} Z^*}$. As a linear time-invariant system $\bm{A}$ is assumed to govern the dynamics between sequential frames, an evolution of a frame $\bm{\psi_t}$ can be formulated according to an initial frame $\bm{\psi_0}$ as below.
\begin{align}
\bm{
\psi_t = A^t \psi_0 \approx (U F_{dmd} U^*)^t \psi_0 
	   = U Y D_{\mu}^t Z^* U^* \psi_0}
\end{align}
Let $\bm{\Phi=UY}$ be the DMD modes and $ \bm{A=Z^* U^* \psi_0} $ as the corresponding magnitudes, then a frame $\bm{\psi_t}$ can be rewritten regardless of $\bm{\psi_0}$ as $\bm{\psi_t = \Phi D_{\mu}^t A}$.
With the amplitudes in $\bm{D_{\alpha}}$ and the Vandermonde matrix, $\bm{V_{and}}$, representing temporal evolution and $\bm{\Phi}$ representing spatial modes, the frame cluster $\bm{\Phi_0}$ is reformulated as follows:
\begin{align}
\label{sec-dpm:subsec-dmdsp:subsubsec-dmd:eqn-3} 
\bm{\Psi_0} &= \bm{[\psi_0 \ldots \psi_{N-1}] \approx \Phi D_{\alpha} V_{and}} \\ \nonumber
	&= \bm{[\phi_0 \ldots \phi_r] }
    \left[
    \begin{matrix}
    	\alpha_1 & 	& \\
        	  & \ddots & \\
              & 	& \alpha_r
    \end{matrix}    
    \right]
    \left[
      \begin{matrix}
          1 & \ldots & \mu_1^{N-1} \\
          \vdots & \ddots & \vdots \\
          1      & 	& \mu_r^{N-1}
      \end{matrix}            
    \right]       
\end{align}
Determination of unknown amplitudes $\bm{\alpha}:=[\alpha_1 \ldots \alpha_r]^T$ depends on the solution of the following optimization problem.
\begin{align}
\argmin_{\bm{\alpha}} \|\bm{J(\alpha)}\}| = \argmin_{\bm{\alpha}} \| \bm {\Psi_0 - \Phi D_\alpha V_{and}} \|_F^2
\label{sec-dpm:subsec-dmdsp:subsubsec-dmd:eqn-4}
\end{align}
where the optimal DMD amplitudes can be obtained by, 
\begin{align}
\bm{\alpha_{dmd}} 
= ((\bm{Y^* Y}) \circ (\bar{\bm{V_{and} V^*_{and}))^{-1}}} diag(\bm{V_{and} V \Sigma^*Y})
\label{sec-dpm:subsec-dmdsp:subsubsec-dmd:eqn-5}
\end{align}

\subsubsection{Sparsity Promoting Dynamic Mode Decomposition}
\label{sec-dpm:subsec-dmdsp:subsubsec-dmdsp}
Though an optimal amplitude for each mode is found by the Equation (\ref{sec-dpm:subsec-dmdsp:subsubsec-dmd:eqn-4}), the amplitudes are solved with the assumption that every mode is equally significant. However, it is not always true and especially not correct for dynamics of micro-expressions. These dynamics are sparse as these expressions are naturally \change[CL]{short}{concise} and sudden. Therefore, only these sparse modes are significant while the rest can be removed without much loss in re-construction of the original signals. In order to reveal the sparsity of the dynamics and its approprate amplitudes, we adopt the sparsity-promoting DMD approach of Jovanovic et al. \cite{jovanovic2014sparsity} which adds sparse constraints into the DMD formulation. It allows trade-off between loss of signal reconstruction and the number of sparse modes.
\note[RP]{Raph asks: can be inspired from this formulation in our aiming to encode sparsity into LDA?}

As DMD only analyzes data and does not apply any sparse constraints, the number of DMD modes\remove[CL]{as well as the number of reconstructed frames} is equal to the number of frames. DMDSP \change[CL]{allows selection of}{selects} a subset of these DMD modes which have dominant influence on the reconstruction of a given sequence. Implementation of sparsity-constraints involves the following two steps.
\begin{enumerate}
\item Identification of sparsity structure such that a user-defined trade-off between the number of extracted modes and approximation error is achieved.
\item Identification of optimal amplitudes for extracted modes given the sparsity structure.
\end{enumerate}
Jovanovic et al. \cite{jovanovic2014sparsity} suggests that the sparse structure problem in the first step can be relaxed and formulated as the $l_1$-norm of the vector of magnitudes $\bm{\alpha}$:
\begin{align}
\argmin_{\bm{\alpha}} J(\bm{\alpha}) + \gamma \sum_{i=1}^{r} | \alpha_i |
\label{sec-dpm:subsec-dmdsp:subsubsec-dmdsp:eqn-1}
\end{align}
where $\gamma$ is a sparsity regularization parameter and $\alpha_i \in \bm{\alpha}$ is a DMD magnitude at rank-i\change[CL]{. which shows }{, showing} sparseness of the vector $\bm{\alpha}$. The Alternative Direction Method of Multipliers (ADMM) method \cite{wen2010alternating} is utilized to solve the above convex optimization problem. Given the fixed sparsity structure, vectors of magnitudes $\alpha$ can be optimized as a solution to the following constrained convex optimization problem:
\begin{align}
\argmin_{\alpha} \left( | \bm{J(\alpha)} | \right) \quad \text{ s.t. } \quad \bm{E^T\alpha} = 0
\label{sec-dpm:subsec-dmdsp:subsubsec-dmdsp:eqn-2}
\end{align}
where the matrix $E$ represents the sparsity structure of the amplitude vector $\bm{\alpha} \in R^{r \times m}$, identified in the first step; $m$ represents the number of $\alpha_i$ with zero values. Each column vector has only one non-zero element corresponding to \add[CL]{each} zero component\remove[CL]{s} of $\bm{\alpha}$; for example, with $\bm{\alpha} \in \mathbb{C}^4$ and $\bm{\alpha} = [\alpha_1,0,\alpha_3,0]^T$, $E$ is given as follows.
\begin{align}
	\bm{E^T} = \left[ 
				\begin{matrix}
					0 & 1 & 0 & 0 \\
					0 & 0 & 0 & 1 
				\end{matrix}
	\label{sec-dpm:subsec-dmdsp:subsubsec-dmdsp:eqn-4}				
\right]
\end{align}
 Jovanovic et al. \cite{jovanovic2014sparsity} show\remove[CL]{ed} that the optimal DMD amplitudes with a fixed sparsity structure can be computed as follows.
\begin{align}
	\bm{\alpha_{dmdsp}} = 	
	\left[
    	\begin{matrix}
        \bm{I} & 0
        \end{matrix}
    \right]
	\left[ 
    	\begin{matrix} 
        	\bm{P} & \bm{E} \\
           	\bm{E^T} & \bm{0}
        \end{matrix}
    \right]^{-1}
    \left[
    \begin{matrix}
    	\bm{q} \\ \bm{0}
    \end{matrix}
    \right]
\label{sec-dpm:subsec-dmdsp:subsubsec-dmdsp:eqn-3}
\end{align}
More details about solutions of these sparsity-constrained problems can be found in \cite{jovanovic2014sparsity}.

\subsection{Sparse \& Uniform Sampling for Redundancy Removal}
\label{sec-dpm:subsec-gsa}
In the uniform approach, TIM \add[CL]{partially} removes \remove[CL]{a part of the} redundant information by interpolating or synthesizing fewer frames at an arbitrarily temporal grid. This new grid is defined by the number of interpolated frames such that they are temporally equispaced. This technique of redundancy removal only works if more frames are interpolated at the significant part and less frames are interpolated at the insignificant part of a sample signal. For example, the discrete signal of the Figure \ref{sec-dpm:fig-1} has 24 points nine of which are in a significant region and the rest are not. When the uniform sampling interpolates the signal and synthesizes only 13 points in which six are in significant regions and the rest are not\change[CL]{, t}{. T}he ratio between redundant and significant points is dropped from $\frac{15}{9}$ of the original signal to $\frac{7}{6}$ of the synthesized signal. It means that the generated signal becomes less redundant. The significant drawback of this approach is that it interpolates new frames at regularly separated positions regardless of the signal's sparse structure. As positions of interpolated frames on the signal are decided by how many frames should be interpolated, that number can only be arbitrarily decided without \add{any} knowledge about structures of input signals. Moreover, even with this prior knowledge, finding optimal numbers of generated frames is also difficult due to regularly spaced temporal interpolation. The interpolation is unadaptive to the signals' structures. In brief, there is no specific method to guarantee optimal removals of redundancy \add[CL]{in the case of uniform sampling}.

The sparse sampling approach, DMDSP, tackles significant drawbacks in the uniform sampling approach, TIM. While DMD, like TIM, only analyzes the signal and models dynamics in a sample signal regardless of its sparse structure, DMDSP (sparse-promoting DMD) has incorporated the sparse constraints into the analysis\change[CL]{, which are}{. It is} formulated as a convex optimization problem in Equations (\ref{sec-dpm:subsec-dmdsp:subsubsec-dmdsp:eqn-2}) and (\ref{sec-dpm:subsec-dmdsp:subsubsec-dmdsp:eqn-3}). Solutions of these problems are sparse structures $E$ of the sample signal and their optimal amplitudes $\alpha$. With the sparsity constraint, amplitudes of dynamics are large if they have profound contribution on the approximate reconstruction of original sequences. Otherwise, their amplitudes are small or nearly zero. In \change[CL]{a}{the} sparse sampling approach, redundancy is eliminated by removing modes with small or near-zero amplitudes; then, sequences are reconstructed with the \change[CL]{rest}{remaining dominant modes}. The \change[CL]{reconstructed}{reassembled} signal is shorter and only contains significant parts of the signal, as demonstrated in Figure \ref{sec-dpm:subsec-dmdsp:fig-1}, a result of applying sparse sampling on the sample signal of Figure \ref{sec-dpm:fig-1}. The reconstructed signal in Figure \ref{sec-dpm:subsec-dmdsp:fig-1} demonstrates the main advantage of sparse sampling over the uniform one as redundancy of the dynamics in a signal is removed more accurately 
by DMDSP than TIM.

\section{Dynamic Analysis of Spontaneous Subtle Emotions}
\label{sec-dan}
Since human beings control several muscles, either consciously or sub-consciously, beneath their skins to form facial expressions as the main non-verbal communication of their emotional states, characteristics of the expressions heavily depend on muscular movements. Understandably, researches of automatic facial expression recognition show superiority of spatiotemporal features over their spatial counterparts i.e. LBPTOP has been shown to be better than LBP and Gabor-like features in \cite{zhao2007dynamic}. It shows that temporal changes or dynamics across consecutive frames significantly contribute to recognition of normal facial expressions. For subtle expressions, careful analysis of temporal dynamics becomes more crucial as subtle facial expressions are not visually far from neutral faces, \change[CL]{at least}{especially} to human eyes. The possibility of correctly recognizing or detecting subtle emotions greatly depends on how discriminating the temporal dynamics across consecutive frames are. Despite their importance, a spatiotemporal feature like LBPTOP only partly utilizes these dynamics without further analysis, visualization or enhancement. To better leverage these subtle facial movements, we utilise the DMD analysis technique, introduced in \ref{sec-dpm:subsec-dmdsp:subsubsec-dmd}, to separate, analyze and visualize facial dynamics. As TIM and DMDSP are only sampling the original sequence uniformly or sparsely, respectively, the result of these pre-processes are sequences of generated frames in either case. Therefore, like original frames, each generated frame can be then analyzed by DMD for their contributions to \remove[CL]{the} overall dynamics regardless of 
sampling \change[CL]{strategy}{strategies}. DMD decomposes an image sequence into three main components: spatial modes $\Phi$, amplitudes $\alpha$ and temporal dynamics $\mu$ as per Equation (\ref{sec-dpm:subsec-dmdsp:subsubsec-dmd:eqn-3}). While spatial modes and temporal dynamics represent separated spatial and temporal information, an amplitude measures a contribution of each spatial mode and its corresponding temporal dynamic of each frame in the reconstruction of the original image sequence. 
Since frames are assumed to be linearly independent, the number of dynamic modes and magnitudes, i.e. the rank $r$, is equal to the number of input frames. As a result, each frame has its corresponding DMD amplitude.\remove[CL]{; it is the amount of contribution from each frame for reconstructing dynamics of original sequences. In other words,} The more activities each frame of facial expressions has, \remove[CL]{or the more it is different from the rest} the larger its DMD magnitude is. DMD magnitude is a frame-by-frame measurement of facial activities; therefore, we can plot the amplitudes against the frame index as a visualization of micro-expressions' dynamics. These plots \change[CL]{represent the}{show} temporal analyses of dynamics for video sequences of subtle emotions \change[CL]{, i.e. in our case}{from} the CASME II and SMIC corpora. A disadvantage of the above visualization method is that this temporal analysis is only applicable for individual video sequences. However, it is not able to be generalized for the whole databases collectively due to the variety of frame-lengths among video sequences. To visualize the general dynamics of subtle emotions in a database, we propose spectral analyses in place of temporal analyses as the frequency bandwidth of dynamics is fixed to a certain range with respect to sampling rates or recording frame rates regardless of frame-lengths. With temporal and spectral analyses, we can visualize and analyze the dynamics of both individual samples and whole databases. When redundancy of dynamics is gradually removed, these analyses provide observations of changes in dynamics at both levels. Therefore, they help identify which temporal parts and frequency bandwidths are important or redundant for a video sequence of a micro-expression or a whole database of spontaneous subtle emotions. These observations of significant temporal and spectral parts of signals can be compared with prior knowledge about micro-expressions such as the duration of micro-expressions lasting between $\frac{1}{25}$s and $\frac{1}{15}$s \cite{ekmanMicro}. More importantly, they provide visualization of dynamic behaviors for several classes of subtle expressions. Hence, we can learn about temporal and spectral dynamical characteristics of each type of subtle emotion across different levels of redundancy removals or dynamic sparsity.

\subsection{Temporal Analysis}
\label{sec-dyn:subsec-tem}
Analysis of dynamics in the temporal domain \change[CL]{involves} involves plotting the DMD amplitudes of an image sequence against corresponding frame indexes. The plots of amplitudes over time give a hint at temporal locations where most motions happen. In this section, the temporal analyses are done for sequences\change[CL]{which are}{,} pre-processed by uniform and sparse sampling approaches of redundancy removals for several percentages of preserved frames. They help analyze and visualize how sparse and uniform sampling affect overall dynamics of sequences. The following subsections demonstrate without loss of generality the temporal analysis of uniform and sparse dynamics for two randomly selected samples from the CASME II database ( EP02\_01f of Subject 01, EP08\_03 of Subject 17 ).

\subsubsection{Temporal Analysis of Uniform Sampling}
\label{sec-dyn:subsec-tem-gen}
\begin{figure}
	\IfFileExists{temporalAnalysisOnCASME2_Sub01_EP02_01f_timeTRF_generative_plot.tex}
	{\input{temporalAnalysisOnCASME2_Sub01_EP02_01f_timeTRF_generative_plot.tex}}
	{\includegraphics{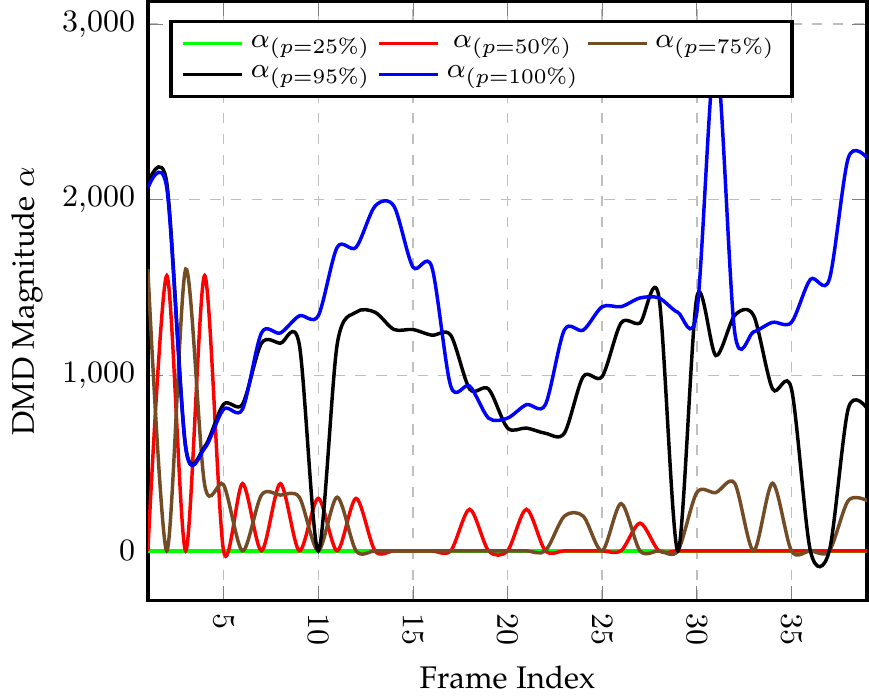}}
	\captionof{figure}{Temporal Analysis with Uniform Sampling in which plots illustrate facial dynamics of Subject 01 in the sample EP02\_01f of the CASME II corpus w.r.t. percentages ($p$) of preserved frames: 25\%, 50\%, 75\%, 95\% and 100\% of the original frame length}
	\label{sec-dyn:subsec-tem-gen:fig-1}
\end{figure}

\begin{figure}
	\IfFileExists{temporalAnalysisOnCASME2_Sub17_EP08_03_timeTRF_generative_plot.tex}
	{\input{temporalAnalysisOnCASME2_Sub17_EP08_03_timeTRF_generative_plot.tex}}
	{\includegraphics{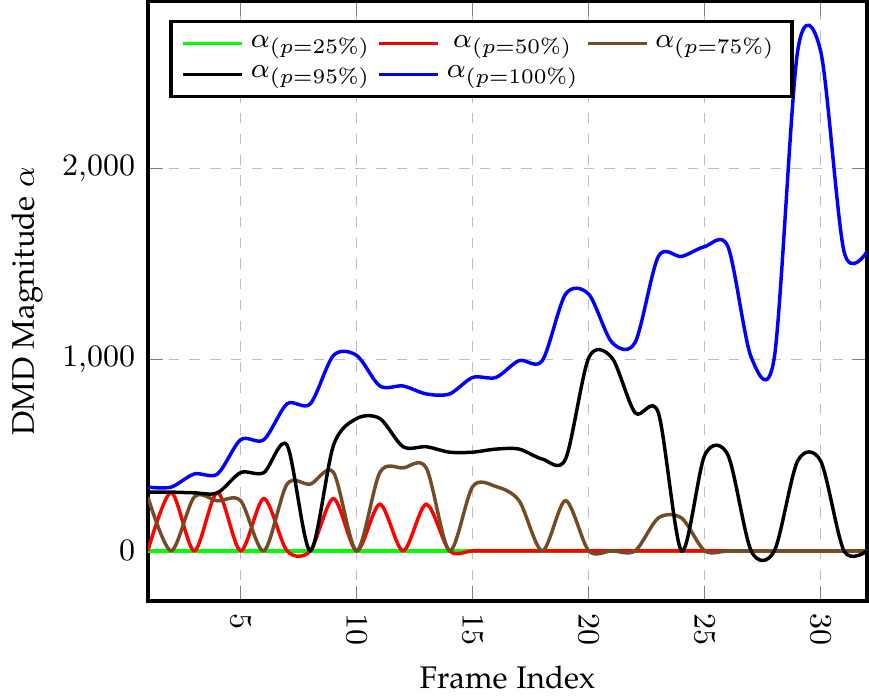}}
	\captionof{figure}{Temporal Analysis with Uniform Sampling in which plots illustrate facial dynamics of Subject 17 in the sample EP08\_03 of the CASME II corpus w.r.t percentages ($p$) of preserved frames: 25\%, 50\%, 75\%, 95\% and 100\% of the original frame length}
	\label{sec-dyn:subsec-tem-gen:fig-2}
\end{figure}

In the uniform sampling, dynamics are learned through interpolation of an original sequence and regenerated into a shorter one. This generated sequence is then analyzed by \change[CL]{Dynamic Mode Decomposition}{DMD} for the dynamic amplitudes of corresponding frames. As TIM regularly samples \change[CL]{the}{an} original dynamic by a temporally equispaced grid, the DMD amplitudes 
can be mapped to the original sequence in the same grid. For example, lets assume an input sequence of 32 frames; TIM interpolates five frames at the following indexes: 1,8,16,24,32. Then DMD amplitudes of these five frames correspond to dynamics at five positions 1,8,16,24,32 of the original sequence while other positions are supposed to have zero-dynamics. \change[CL]{In addition, the number of frames remaining after TIM interpolation is used for computing how much percentage of the original frame-lengths is preserved}{The number of interpolated frames over the original frame length is a percentage of preservation}. The above approach is used for computing and plotting temporal dynamics of the chosen sequences in Figures \ref{sec-dyn:subsec-tem-gen:fig-1}, \ref{sec-dyn:subsec-tem-gen:fig-2}. In these plots, 25\%, 50\%, 75\%, 95\% and 100\% of the frames are preserved for temporal analysis. Note that 100\% means no TIM interpolation of original frames has been done \change[CL]{, which}{and a video sequence} is directly analyzed by DMD for dynamical amplitudes. As TIM interpolation likely down-samples 
the dynamics, the temporal dynamics of lower percentages of preserved frames look like the blurred versions of higher percentages.

\subsubsection{Temporal Analysis of Sparse Sampling}
\label{sec-dyn:subsec-tem-sel}
\begin{figure}
	\IfFileExists{temporalAnalysisOnCASME2_Sub01_EP02_01f_timeTRF_selective_plot.tex}
	{\input{temporalAnalysisOnCASME2_Sub01_EP02_01f_timeTRF_selective_plot.tex}}
	{\includegraphics{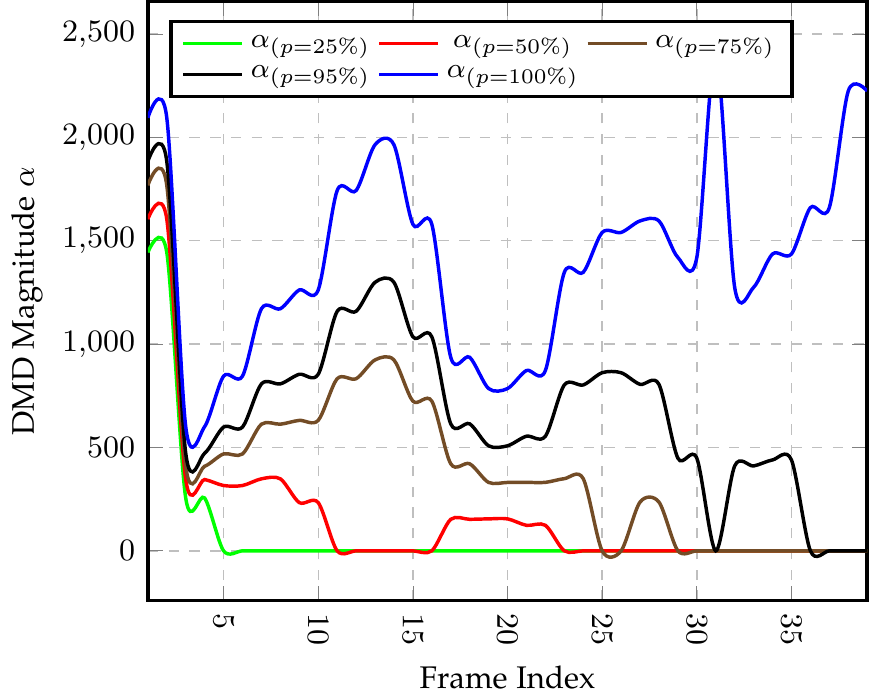}}
	\caption{Temporal Analysis with Sparse Sampling in which plots illustrate facial dynamics of Subject 01 in the sample EP02\_01f of the CASME II corpus w.r.t percentages of preserved frames: 25\%, 50\%, 75\%, 95\% and 100\% of the original frame length}
	\label{sec-dyn:subsec-tem-sel:fig-1}
\end{figure}

\begin{figure}
	\IfFileExists{temporalAnalysisOnCASME2_Sub17_EP08_03_timeTRF_selective_plot.tex}
	{\input{temporalAnalysisOnCASME2_Sub17_EP08_03_timeTRF_selective_plot.tex}}
	{\includegraphics{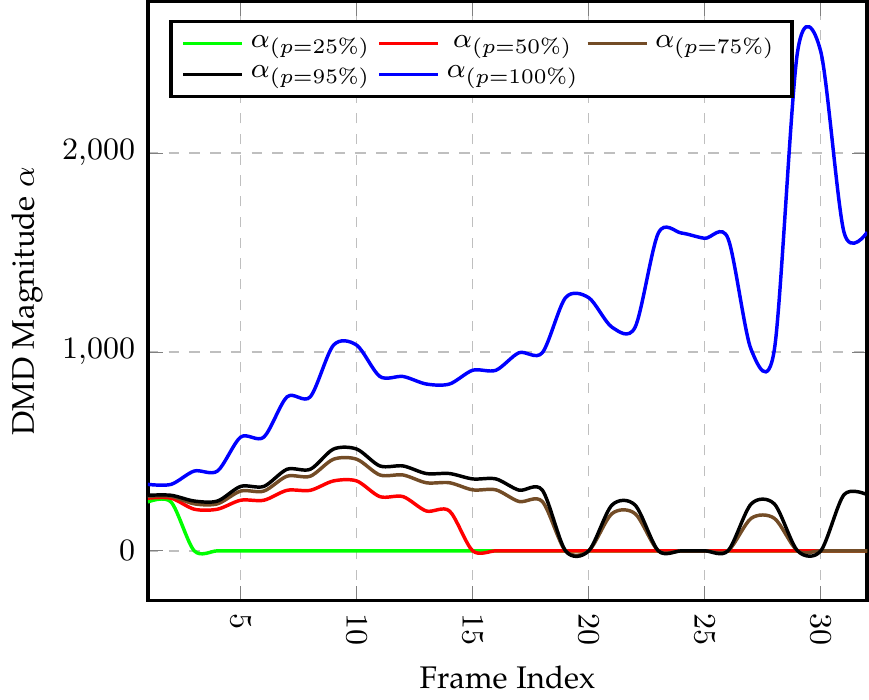}}
	\caption{Temporal Analysis with Sparse Sampling in which plots illustrate facial dynamics of Subject 17 in the sample EP08\_03 of the CASME II corpus w.r.t percentages of preserved frames: 25\%, 50\%, 75\%, 95\% and 100\% of the original frame length}
	\label{sec-dyn:subsec-tem-sel:fig-2}
\end{figure}

As the sparse sampling approach \change[CL]{learns}{looks for} significant frames of original dynamics, it is possible to determine redundant frames among original sequences. Correspondingly, amplitudes of these frames are masked to zeros while the rest of the amplitudes are analyzed in DMD. Since sparse structures are regulated by the $\gamma$ parameter, its value also controls how many percentages of frames are preserved. For a certain range of $\gamma$
value $[38-20000]$, these percentages vary between 5\% and 100\%. Figures \ref{sec-dyn:subsec-tem-sel:fig-1} and \ref{sec-dyn:subsec-tem-sel:fig-2} show temporal analyses of sparsely sampled dynamics for 25\%, 50\%, 75\%, 95\% and 100\% preserved frames. Note that 100\% means the DMD amplitudes of original sequences are not processed by DMDSP. These plots show how the sparse dynamics change across different percentages\change[CL]{ and}{. Furthermore,} they show consistency of sparse structures, learned by DMDSP with different $\gamma$ values, especially in Figure \ref{sec-dyn:subsec-tem-sel:fig-2}. 

Comparisons between temporal analyses of uniform and sparse sampling \remove[CL]{of the sample samples} ( Figure \ref{sec-dyn:subsec-tem-gen:fig-1} vs Figure \ref{sec-dyn:subsec-tem-sel:fig-1} or Figure \ref{sec-dyn:subsec-tem-gen:fig-2} vs Figure \ref{sec-dyn:subsec-tem-sel:fig-2} ) show that sparse dynamics have more consistent and clearer patterns than uniform dynamics. For instance, plots of DMD magnitudes of the sample EP08\_03 
have consistent shapes for frames 0-19 across multiple percentages (25\%, 50\%, 75\%) of preservation. Meanwhile, uniform sampling blindly selects equispaced indexes and removes in-between frames. Therefore, plots of DMD magnitudes $\alpha$ are radically different given various percentages of preservation $p$ in the Figure \ref{sec-dyn:subsec-tem-gen:fig-2}. 
Furthermore, the DMD magnitudes $\alpha$ in the Figure \ref{sec-dyn:subsec-tem-gen:fig-2} represent how much each frame contributes into the reconstruction of original dynamics. Therefore, consistent structures of sparse sampling plots, Fig. \ref{sec-dyn:subsec-tem-sel:fig-2}, and their magnitudes provide locations and weights of significant dynamics in an image sequence.

\subsection{Spectral Analysis}
\label{sec-dyn:subsec-spa}
\begin{figure}
\IfFileExists{spectralAnalysisOnCASME2_freqTRF_plot.tex}
{\input{spectralAnalysisOnCASME2_freqTRF_plot.tex}}
{\includegraphics{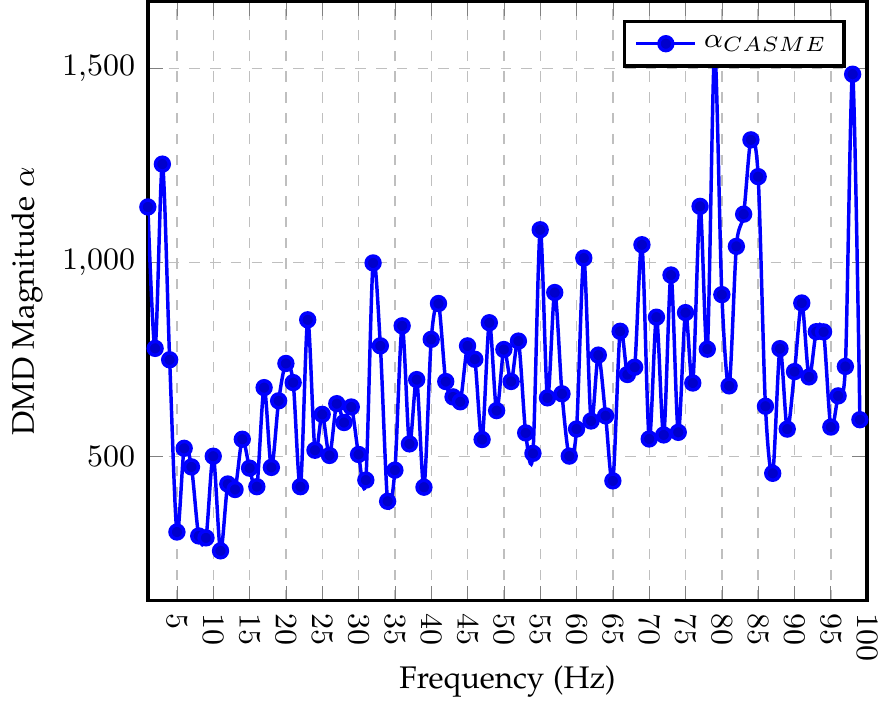}}
\caption{Spectral Analysis on Spontaneous Subtle Emotions of CASME II}
\label{sec-dyn:subsec-spa:fig-1}
\end{figure}

\begin{figure}
\IfFileExists{spectralAnalysisOnSMIC_freqTRF_plot.tex}
{\input{spectralAnalysisOnSMIC_freqTRF_plot.tex}}
{\includegraphics{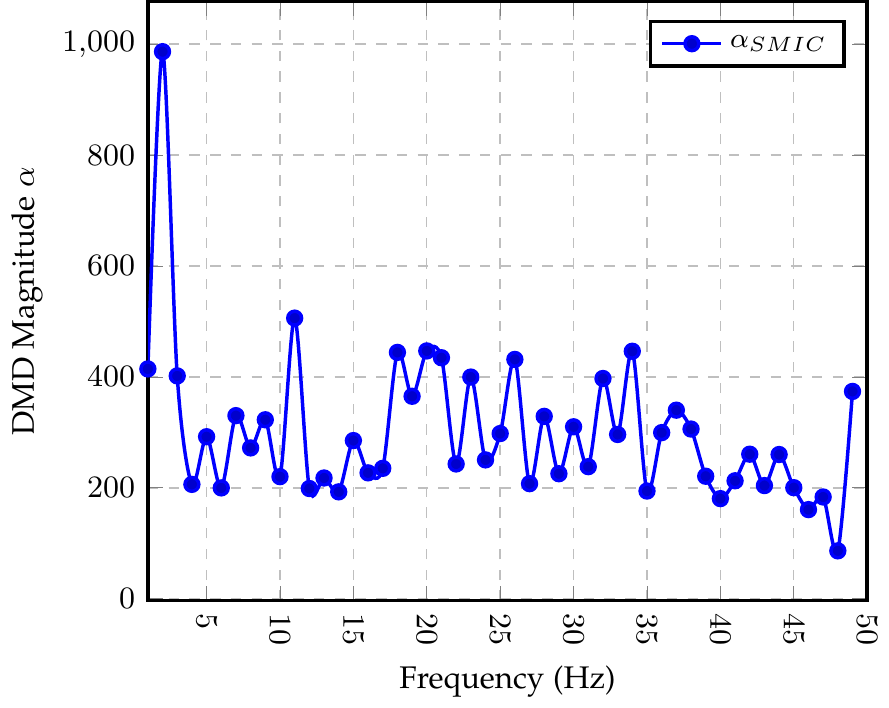}}
\caption{Spectral Analysis on Spontaneous Subtle Emotions of SMIC}
\label{sec-dyn:subsec-spa:fig-2}
\end{figure}

Temporal analysis is only useful for inspecting dynamics of a single sample. It is not able to summarize and visualize dynamics of several samples due to varying frame-lengths across different samples. Therefore, temporal analysis is unable to show dynamical characteristics of the whole CASME II or SMIC databases. Spectral analysis avoids the frame-length problem through analyzing video samples of databases in the frequency domain. As DMD decomposes an image sequence into spatial modes $\phi$, amplitudes $\alpha$ and temporal dynamics $\mu$ in Equation (\ref{sec-dpm:subsec-dmdsp:subsubsec-dmd:eqn-3}), the frequency of dynamics $f_{DMD}$ can be computed from the temporal dynamics $\mu$ as follows.
\begin{align}
f_{DMD} = \log(\mu)f_s
\label{sec-dyn:subsec-spa:eqn-1}
\end{align}
where $f_s$ is the maximum sampling frequency or frame-rate of an image sequence. The frame-rate also limits the bandwidth of dynamics as $f_{DMD} \in B_{DMD} = [0 \ldots f_s/2]$ Hz according to the Nyquist-Shannon Sampling Theorem \cite{shan2009facial} which states that the maximum frequency of a signal is half of its sampling frequency. For example, the CASME II database is recorded with frame-rate 200 fps; then, its spectral bandwidth of dynamics is $[0-100]$ Hz. Similarly, SMIC has $[0-50]$ Hz bandwidth according to its frame-rate 100 fps. With the Equation (\ref{sec-dyn:subsec-spa:eqn-1}), a frequency value $f_{DMD}$ can be found for each temporal dynamic value $\mu$; therefore, each $f_{DMD}$ has a corresponding amplitude $\alpha$. Moreover, spectral analysis of an image sequence results in a histogram of amplitudes $\alpha$ over bins of frequencies $f_{DMD}$. Given that all videos in the same database have the same frame-rate, bandwidth, hence the same number of bins, it is valid to sum up histograms of all sequences from the same database to produce spectral analysis of a database. For instance, Figure \ref{sec-dyn:subsec-spa:fig-1} shows the spectral analysis with DMD of CASME II and Figure \ref{sec-dyn:subsec-spa:fig-2} demonstrates the dynamics of SMIC over its spectral bandwidth. These figures show the spectra of temporal dynamics from original videos of CASME II and SMIC, which are not pre-processed by neither uniform nor sparse sampling. These spectra of both databases are spread over the whole bandwidth; therefore they do not provide any information about dominant temporal dynamics of micro-expressions. The lack of dominant spectra might be due to noises, generated by redundancies of neutral faces as their differences are not due to motions but illuminating conditions.

The proposed sparse sampling techniques remove redundant neutral frames and reveal significant spectra of dynamics. In addition, plots of spectral analysis  visualize the dominant spectra and how a whole database responds to different sparsity levels or percentages of preserved frames. In the following subsections, spectral analyses are used for analyzing uniformly and sparsely sampling approaches with respect to different percentages. \remove[CL]{of preserved frames.} \change[CL]{It noted}{Note that} these percentages are inversely proportional to the sparsity levels. Five percentage values \add[CL]{i.e.} 5\%, 25\%, 50\%, 75\% and 100\% of preserved frames over the original frame-lengths are used in the following analysis of both uniform and sparse sampling.

\subsubsection{Spectral Analysis of Uniformly Sampling Dynamics}
\label{sec-dyn:subsec-spa-gen}
\begin{figure}
\IfFileExists{spectralAnalysisOnCASME2_freqTRF_generative_plot.tex}
{\input{spectralAnalysisOnCASME2_freqTRF_generative_plot.tex}}
{\includegraphics{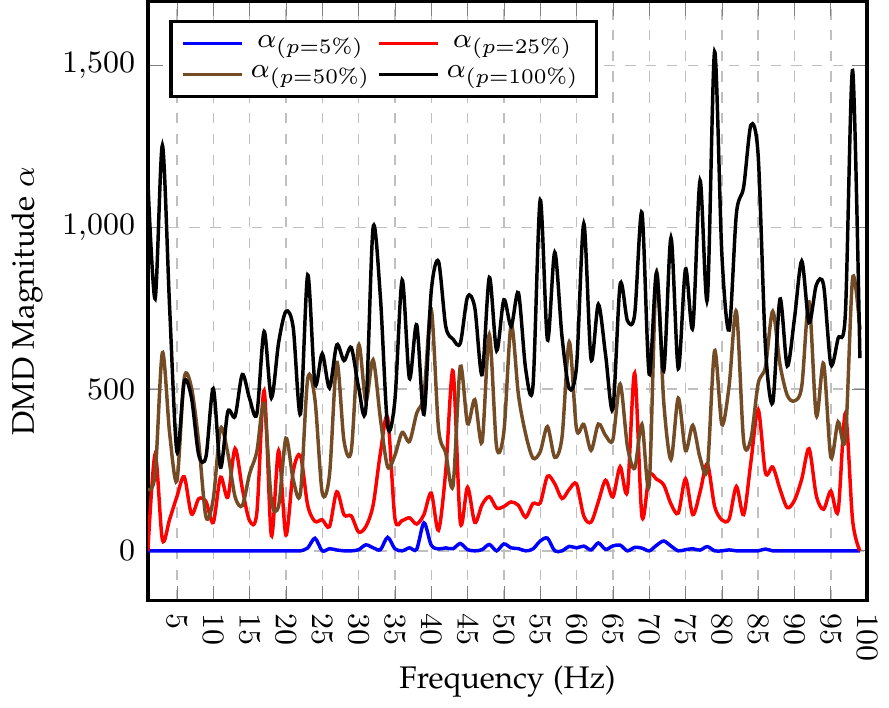}}
\caption{Spectral Analysis of CASME II database pre-processed with Uniformly Sampling approach and four percentages of preserved frames $p=\{5\%,25\%,50\%,100\%\}$}
\label{sec-dyn:subsec-spa-gen:fig-1}
\end{figure}

\begin{figure}
\IfFileExists{spectralAnalysisOnSMIC_freqTRF_generative_plot.tex}
{\input{spectralAnalysisOnSMIC_freqTRF_generative_plot.tex}}
{\includegraphics{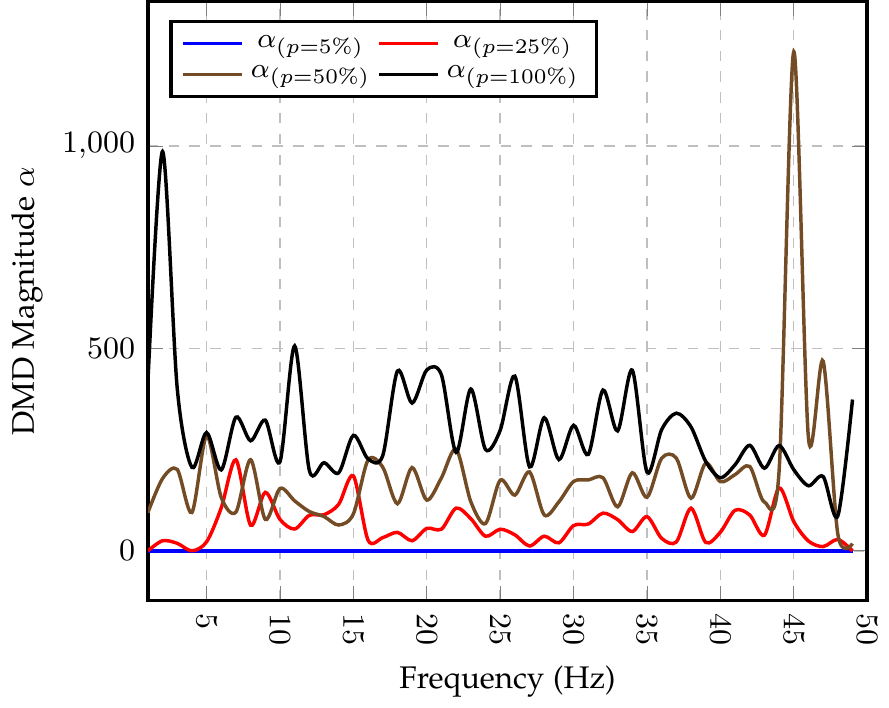}}
\caption{Spectral Analysis of SMIC database pre-processed by Uniformly Sampling approach and four percentages of preserved frames $p=\{5\%,25\%,50\%,100\%\}$}
\label{sec-dyn:subsec-spa-gen:fig-2}
\end{figure}

Section \ref{sec-dpm:subsec-gsa} discussed the technicality of the uniform sampling approach like TIM \cite{pfister2011recognising} for removing dynamic redundancies. 
In this section, we visualize redundancy removals of the uniform sampling in the spectral domain for whole databases. Figures \ref{sec-dyn:subsec-spa-gen:fig-1} and \ref{sec-dyn:subsec-spa-gen:fig-2} demonstrate several spectral analyses of CASME II and SMIC with respect to four 
percentages of preserved frames $p = \{5\%, 25\%, 50\%, 100\%\}$, four plots with different colors. These plots in Figures \ref{sec-dyn:subsec-spa-gen:fig-1} and \ref{sec-dyn:subsec-spa-gen:fig-2} show that the uniform sampling approach tends to suppress the dynamics of the lowest and highest parts of the bandwidth while they smoothen the dynamics in the middle frequency ranges. In other words, the uniform sampling approach behaves like a band-pass filter for dynamics since frames are interpolated at equispaced points along a manifold.  
It linearly combines nearby original frames, preserves spatial characteristics whilst smoothing temporal profiles \cite{pfister2011recognising}\cite{zhou2011towards}. 
 

\subsubsection{Spectral Analysis of Sparsely Sampling Dynamics}
\label{sec-dyn:subsec-spa-sel}

In contrast to the previous section \ref{sec-dyn:subsec-spa-gen} about spectral analyses for uniform dynamics by TIM \cite{pfister2011recognising} on CASME II and SMIC databases, this section shows spectral analysis of sparse dynamics with DMDSP \cite{jovanovic2014sparsity}. While Section \ref{sec-dpm:subsec-dmdsp:subsubsec-dmdsp} elaborates how the sparsity constraint with \change[CL]{the}{a} regulation parameter $\gamma$ is formulated in DMDSP, this section explains a relation between sparsity levels and percentages of preserved frames. More importantly, it \change[CL]{provides sparsity analysis of}{analyses} the sparse dynamics on CASME II and SMIC databases.

\begin{figure}
\IfFileExists{gammaAnalysis_gammaTRF_plot.tex}
{\input{gammaAnalysis_gammaTRF_plot.tex}}
{\includegraphics{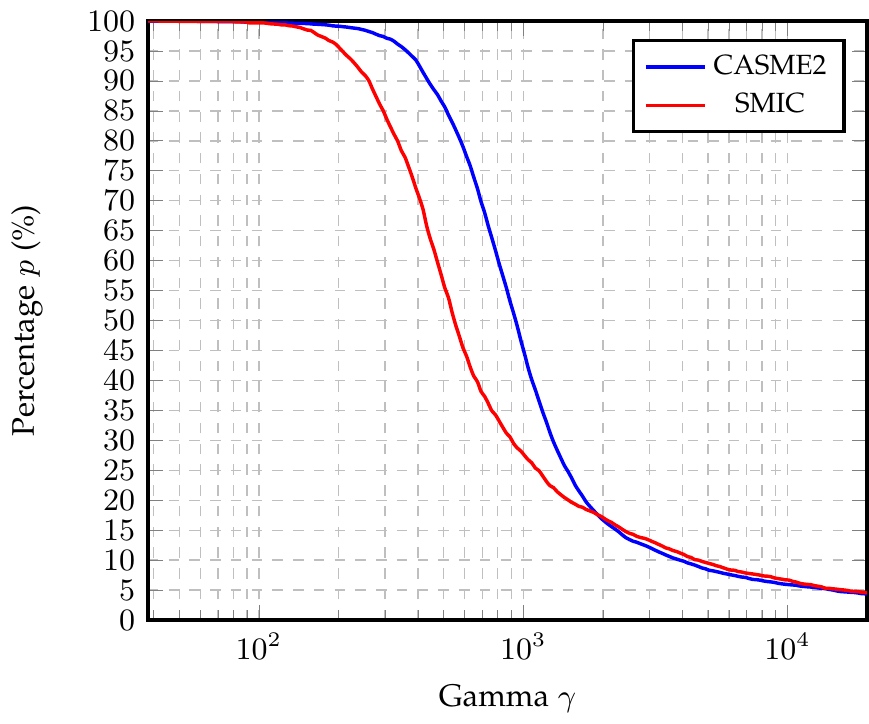}}
\caption{Gamma Analysis of CASME II and SMIC databases with Sparse Sampling approach, which demonstrates inversely proportional relation between percentages of preserved frames $p$ and sparsity parameter $\gamma$}
\label{sec-dyn:subsec-spa-sel:fig-1}
\end{figure}

\begin{figure}
\IfFileExists{spectralAnalysisOnCASME2_freqTRF_selective_plot.tex}
{\input{spectralAnalysisOnCASME2_freqTRF_selective_plot.tex}}
{\includegraphics{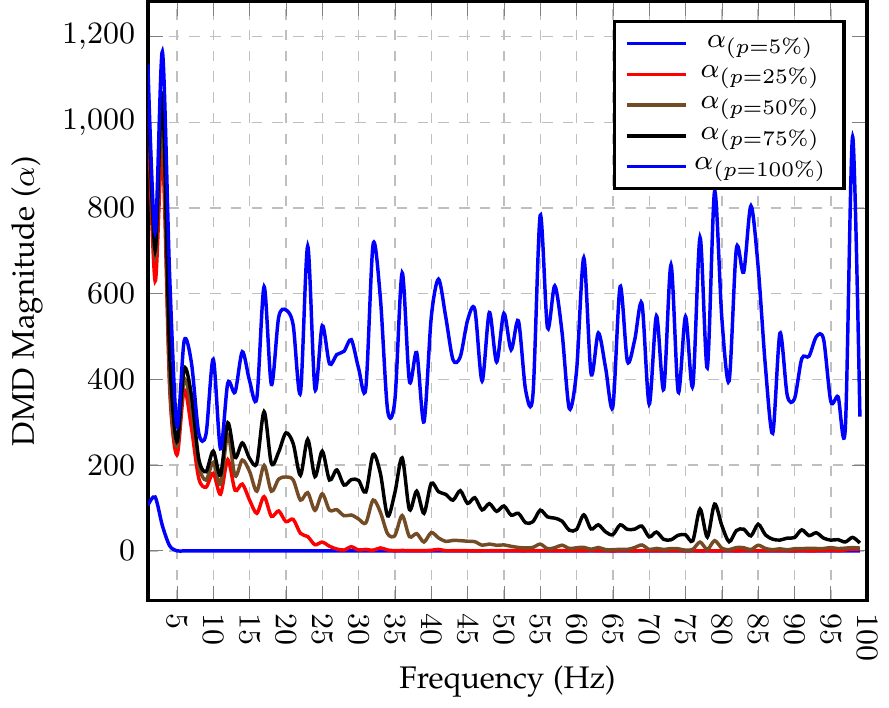}}
\caption{Spectral Analysis of CASME II database, pre-processed by Sparse Sampling approach and five percentages of preserved frames $p=\{5\%,25\%,50\%,75\%,100\}$}
\label{sec-dyn:subsec-spa-sel:fig-2}
\end{figure}

\begin{figure}
\IfFileExists{spectralAnalysisOnSMIC_freqTRF_selective_plot.tex}
{\input{spectralAnalysisOnSMIC_freqTRF_selective_plot.tex}}
{\includegraphics{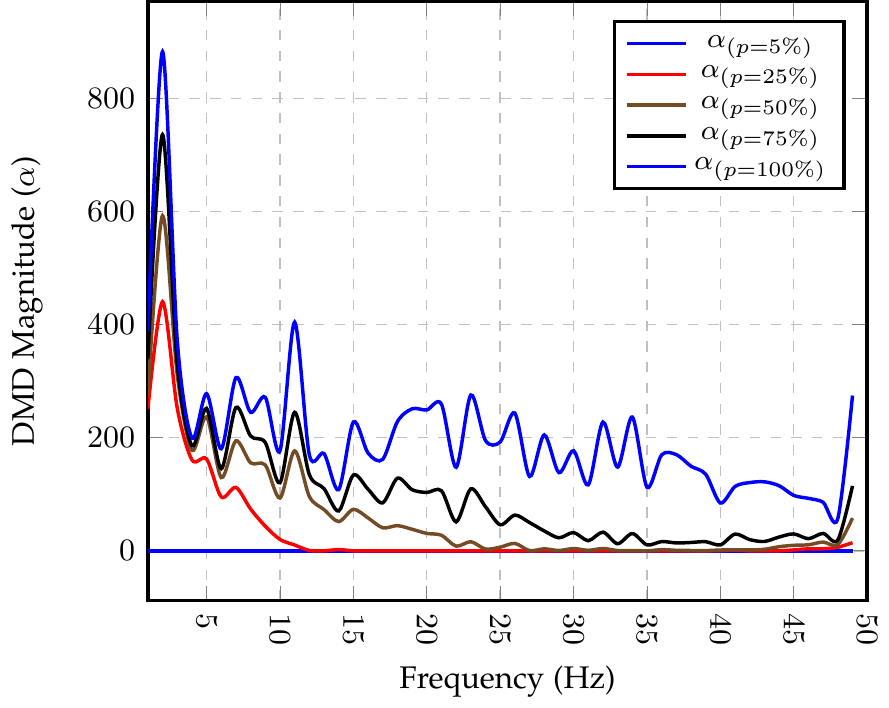}}
\caption{Spectral Analysis of SMIC database, pre-processed by Sparse Sampling approach and five percentages of preserved frames $p=\{5\%,25\%,50\%,75\%,100\%\}$}
\label{sec-dyn:subsec-spa-sel:fig-3}
\end{figure}

The regulation parameter $\gamma$ in Equation (\ref{sec-dpm:subsec-dmdsp:subsubsec-dmdsp:eqn-1}) controls \change[CL]{how sparse DMDSP would tolerate}{sparseness enforced by DMDSP}. The higher the $\gamma$ values are, the more sparsity constraints are enforced. In other words, more dynamics are removed with increments of $\gamma$ values. However, the optimization problem in Equation (\ref{sec-dpm:subsec-dmdsp:subsubsec-dmdsp:eqn-1}) also involves losses during reconstruction from the sparse dynamics. To achieve this optimization, trade-offs are made between reconstruction loss and the number of dynamics. \remove[CL]{The natural solution for this optimization is gradually removing dynamics with decreasing importance.} \change[CL]{It also means increasing removals of}{The solution is to increasingly removing} unimportant frames from the sequence as the sparsity level i.e. $\gamma$ increases. Hence, there is an inversely proportional relation between sparsity values $\gamma$ and percentages of preserved frames. This relationship is shown through plots of 
percentages over logarithmic ranges of $\gamma$ values in Figure \ref{sec-dyn:subsec-spa-sel:fig-1}.
To generate the Figure \ref{sec-dyn:subsec-spa-sel:fig-1}, we choose 400 gamma values, equispaced in the logarithmic scale, in a range [38-20000]. DMDSP is applied to a video sample with these gamma values. Due to the sparsity constraint, many suppressed dynamics with near-zero amplitudes are removed from the sequences. From the number of remaining frames, a percentage of preserved frames can be computed for the corresponding gamma value. 
Figures \ref{sec-dyn:subsec-spa-sel:fig-2} and \ref{sec-dyn:subsec-spa-sel:fig-3} show spectral analyses of CASME II and SMIC databases with different percentages of preserved frames for redundancy removals. Spectral analyses are done for 5\%, 10\%, 15\% until 100\%; however, only a few are displayed for simplicity of the plot. Sparse structures are learned with DMDSP, sparsity-promoting DMD, on dynamics of image sequences in the sparse sampling while the uniform sampling with TIM just averagely interpolates from original frames. \change[CL]{In s}{S}pectral analyses of both databases \change[CL]{, we can notice}{show} similar trends in which high-frequency components are suppressed by the decreasing percentages or increments of sparsity. In other words, most sparse structures or dynamically significant frames are located at low frequencies from 0 Hz to 25 Hz. According to Ekman \cite{ekmanMicro}, the duration of subtle emotions is more than $\frac{1}{25}$ s. Therefore, the spectra of temporal dynamics should be significant below 25 Hz, which are well spotted by the sparse sampling approach in both CASME II and SMIC databases.

\section{Experiments \& Discussion}
\label{sec-eds}
\begin{table*}[t]
	\centering
	\scriptsize	
	\caption{Performance measures (F1 score (F1), Recall Rate (R) and Precision Rate (P)) of the proposed \textbf{SS} approach, in comparison to the \textbf{US}, \textbf{US*}, (\textbf{RA}) and \textbf{BL} approaches w.r.t each class of CASME II (O,D,H,S,R) and SMIC (P,N,S) corpora}
	\begin{tabular}{lcccccccccccccccccccccccc}
		\toprule
		& \multicolumn{ 15}{c}{\textbf{CASME II}} & \multicolumn{ 9}{c}{\textbf{SMIC}} \\
		\cmidrule(r){2-16} \cmidrule(r){17-25}
		& \multicolumn{ 3}{c}{Others (O)} & \multicolumn{ 3}{c}{Disgust (D)} & \multicolumn{ 3}{c}{Happiness (H)} & \multicolumn{ 3}{c}{Surprise (S)} & \multicolumn{ 3}{c}{Repression (R)} & \multicolumn{ 3}{c}{Negative (N)} & \multicolumn{ 3}{c}{Positive (P)} & \multicolumn{ 3}{c}{Surprise (S)} \\ \cmidrule(r){2-4} \cmidrule(r){5-7} \cmidrule(r){8-10} \cmidrule(r){11-13} \cmidrule(r){14-16} \cmidrule(r){17-19} \cmidrule(r){20-22} \cmidrule(r){23-25}
		& F1 & RR & PR & F1 & RR & PR & F1 & RR & PR & F1 & RR & PR & F1 & RR & PR & F1 & RR & PR & F1 & RR & PR & F1 & RR & PR \\ \midrule
		\textbf{SS} & \textbf{.56} & \textbf{.64} & \textbf{.50} & \textbf{.27} & \textbf{.23} & \textbf{.32} & \textbf{.58} & \textbf{.55} & \textbf{.62} & \textbf{.67} & \textbf{.52} & \textbf{.93} & \textbf{.39} & \textbf{.41} & \textbf{.38} & \textbf{.53} & \textbf{.48} & \textbf{.59} & \textbf{.60} & \textbf{.70} & \textbf{.53} & \textbf{.64} & \textbf{.62} & \textbf{.67} \\
		\textbf{US} & .52 & .58 & .47 & .09 & .07 & .12 & .36 & .48 & .29 & .34 & .24 & .60 & .32 & .30 & .35 & .37 & .36 & .38 & .40 & .46 & .35 & .44 & .37 & .55 \\
		\textbf{US*} & .47 & .52 & .44 & .20 & .19 & .20 & .37 & .44 & .32 & .12 & .08 & .22 & .18 & .15 & .24 & .46 & .41 & .52 & .49 & .53 & .45 & .48 & .51 & .46 \\
		\textbf{RA} & .49 & .54 & .44 & .13 & .12 & .14 & .25 & .42 & .18 & .10 & .06 & .19 & .32 & .30 & .34 & .38 & .36 & .41 & .36 & .46 & .30 & .37 & .29 & .52 \\
		\textbf{BL} & .47 & .53 & .42 & .25 & .22 & .27 & .33 & .31 & .34 & .42 & .40 & .43 & .30 & .26 & .35 & .39 & .36 & .42 & .41 & .49 & .35 & .39 & .35 & .45 \\ \bottomrule
	\end{tabular}
	\label{sec-eds:tab2}
\end{table*}

\begin{table*}[t]
	\begin{minipage}[t]{.5\linewidth}
	\centering		
	\scriptsize
	\caption{Number of video samples for each class \newline of CASME2 (H,D,R,S,O) and SMIC (P,N,S) corpora}	
	\vskip 3pt
	\begin{tabular}{lcclcc}
		\toprule
		\multicolumn{ 3}{c}{\textbf{CASME II}} & \multicolumn{ 3}{c}{\textbf{SMIC}} \\ \cmidrule(r){1-3} \cmidrule(r){4-6}
		Emotion  & Label & \# samples & Emotion & Label & \# samples \\ \midrule
		Happiness & H & 33 & Positive & P & 51 \\ 
		Disgust & D & 60 & Negative & N & 70 \\ 
		Repression & R & 25 & Surprise & S & 43 \\
		Surprise & S & 27 &  &  &  \\
		Others & O & 102 &  &  &  \\ \midrule
		Fear & \textit{N/A} & 2 & & & \\
		Sadness & \textit{N/A} & 7 & & & \\ \bottomrule
	\end{tabular}
	\label{sec-eds:tab3}
	\end{minipage}
	\begin{minipage}[t]{.5\linewidth}
	\centering
	\scriptsize
	\caption{Performance measures (F1 score, Recall Rate and Precision Rate) of the proposed \textbf{SS} approach, in comparison to the \textbf{US}, \textbf{RA} and \textbf{BL} approaches}.	
	\begin{tabular}{l*{8}{c}}
		\toprule
		& \multicolumn{4}{c}{\textbf{CASME II}} & \multicolumn{4}{c}{\textbf{SMIC}} \\
		\cmidrule(r){2-5} \cmidrule(r){6-9}
		& ACC & F1 & RR & PR & ACC & F1 & RR & PR \\
		\midrule
		\textbf{SS} & \textbf{.49} & \textbf{.51} & \textbf{.47} & \textbf{.55} & \textbf{.58} & \textbf{.60} & \textbf{.60} & \textbf{.60} \\
		\textbf{US} & .38 & .35 & .33 & .37 & .40 & .41 & .40 & .43 \\
		\textbf{US*} & .33 & .28 & .27 & .28 & .48 & .48 & .49 & .48 \\
		\textbf{RA} & .34 & .29 & .29 & .29 & .37 & .39 & .37 & .41\\
		\textbf{BL} & .38 & .35 & .34 & .36 & .40 & .40 & .40 & .41 \\
		\bottomrule
	\end{tabular}
	\label{sec-eds:tab1}		
	\end{minipage}
\end{table*}

In this section, our experimental results for different approaches of redundancy removal are reported and analyzed for both the CASME II and SMIC databases. The CASME II corpus \cite{yan2014casme} has samples from seven different classes or labels: "others" (O), "disgust"(D), "happiness" (H), "surprise" (S), "repression" (R), "fear" (F), "sadness" (S). However, only the first five of the mentioned classes have enough samples for any statistically meaningful experiments, as shown in the Table \ref{sec-eds:tab3}, and as considered in the database designers' baseline experiments. Meanwhile, the SMIC corpus includes three different emotional categories: "positive" (P), "negative" (N) and "surprise" (S). Hence, we report experimental results of 5-class and 3-class classification for CASME II and SMIC corpora accordingly. For both CASME II and SMIC, labeling was done by the designers per a video sample rather than per a frame. In other words, all frames of a video sample are assumed to have the same emotional labels\change[CL]{. This is because}{as} each video sample is actually a cropped video clip from \change[CL]{the}{a} raw footage based on detected onset (beginning of the emotion) and offset (ending) points. Hence, the automatic recognition of subtle emotions is trained and tested at the granularity of video samples assuming that facial expressions in a video sample correspond to only one emotional state. The assumption is justified by procedures of sample acquisition in CASME II \cite{yan2014casme} and SMIC \cite{pfister2011recognising}.

\begin{table}[b]
	\scriptsize
	\caption{Experimental parameters of sampling approaches, uniform LBPTOP feature, and Support Vector Machine (SVM) classifier with respect to \textbf{SS}, \textbf{US}, \textbf{US}\textbf{*},\textbf{RA} and \textbf{BL} approaches}
	\begin{center}
		\begin{tabular}{lllcclll}
			\toprule
			& \multicolumn{ 2}{c}{Preprocessing} & \multicolumn{ 2}{c}{Uniform LBPTOP} & \multicolumn{ 3}{c}{SVM} \\ 
			\cmidrule(r){2-3} \cmidrule(r){4-5} \cmidrule(r){6-8}
			& Method & Parameter & CASME II & SMIC & K & \multicolumn{1}{l}{c} & \multicolumn{1}{l}{g} \\ \midrule
			\textbf{SS} & DMDSP & [45:5:100] \% & 5x5x1 & 8x8x1 & RBF & $10^4$ & .5 \\ 
			\textbf{US} & TIM & [45:5:100] \% & 5x5x1 & 8x8x1 & RBF & $10^4$ & .5 \\ 
			\textbf{US*} & TIM & \begin{tabular}{@{}l@{}} CASME II: 150 \\ SMIC: 10 \end{tabular} & 5x5x1 & 8x8x1 & RBF & $10^4$ & .5 \\ 
			\textbf{RA} & RAN & [45:5:100] \% & 5x5x1 & 8x8x1 & RBF & $10^4$ & .5 \\ 
			\textbf{BL} & N/A & N/A & 5x5x1 & 8x8x1 & RBF & $10^4$ & .5 \\ \bottomrule
		\end{tabular}
	\end{center}
	\label{sec-eds:tab4}
\end{table}

In the following experiments, "Sparse Sampling" (\textbf{SS}), "Uniform Sampling" (\textbf{US}), "$\text{Generative}^{*}$" ($\text{\textbf{US}}^{*}$)  and "Random" (\textbf{RA}) approaches for redundancy removals are compared against the "Baseline" (\textbf{BL}) result of subtle emotion recognition tasks. 
Note that the \textbf{US} and $\mathbf{US^*}$ approaches both utilize TIM for generating temporal dynamics but for different frame-length parameters. In the \textbf{US} approach, the number of frames is adaptively synthesized with respect to percentages ($45\%,50\%,\ldots,100\%$) of preserved frames for each sequence. The $\mathbf{US^*}$ approach strictly interpolates each video sample into fixed frame-lengths i.e. 150 frames in Wang et al. \cite{wang2014micro} and 10 frames in Pfister et al. \cite{pfister2011recognising}. Meanwhile, the \textbf{BL} approach shows performances of systems without eliminating dynamic redundancies. In the \textbf{RA} approach, frames are randomly collected from the original sequences such that a certain percentage ($45\%,50\%,\ldots,100\%$) of frame-length is acquired. While the \textbf{SS} and \textbf{US} approaches are discussed in detail and comparatively analyzed in previous sections \ref{sec-dpm:subsec-tim} and \ref{sec-dpm:subsec-dmdsp}, inclusion of the \textbf{RA}, \textbf{BL}, \textbf{US} and \textbf{US*} approaches aims to confirm 
the superiority of the proposed Sparse Sampling (\textbf{SS}) approach over random or no elimination of redundant frames as well as uniform sampling approaches from the \change[CL]{original}{previous} works \cite{yan2014casme,pfister2011recognising}. Experimental parameters are further elaborated for the rest of this section and summarized in Table \ref{sec-eds:tab4}.

After dynamics of signals are manipulated and shorter videos are reconstructed from remaining dynamic modes, all video frames of the same database are spatially normalized as they are resized to a fixed resolution. 
Denote that the resized and shortened videos have $n_{r}$ rows, $n_c$ columns and $n_f$ frames. In the CASME II corpus, $n_r$ and $n_c$ are set to 340 and 280 respectively; meanwhile, $n_r$ and $n_c$ for SMIC are fixed at 170 and 140 respectively. 
Then, a feature vector is extracted from each video sample as follows: 
(i) dividing each image sequence into a $n_b \times n_b \times 1$ matrix of non-overlapping volumetric 3-D blocks with size ($\frac{n_r}{n_b},\frac{n_c}{n_b},\frac{n_f}{1}$) where ($n_b = 5$) for CASME II and $n_b = 8$ for SMIC); (ii) extracting a histogram from each block by uniform 4-connected $\text{LBPTOP}_4,4,4,1,1,3$ \cite{zhao2007dynamic} with $(1,1,3)$ as correspondent horizontal ($R_x$), vertical ($R_y$) and temporal ($R_t$) radii; (iii) concatenating all histograms of $n_b \times n_b$
volumetric cells into the feature vector. The number of blocks and LBPTOP configuration are chosen for replicating experimental results from Table 4 in \cite{yan2014casme} and Table 4 in SMIC \cite{li2013spontaneous}. Though Yan et al. \cite{yan2014casme} proposes that $R_t=4$ gives the best accuracy, we choose $R_t=3$ to satisfy minimum frame-length ($2 \times R_t+1=7$) for LBPTOP feature extraction even if only 45\% of original frames are preserved. Given the mentioned details of LBPTOP feature extraction for CASME II and SMIC, dimensions ($D \times 1$) of feature vectors are computed by $D={n_b}^2 \times 15 \times 3$. \remove[CL]{where} Note that 15 is the dimension of each uniform LBPTOP histogram, 3 is the number of considered planes $XY$,$XT$ and $YT$, and ${n_b}^2$ is the number of blocks. As a result, the dimensions ($D$) of features for CASME II and SMIC experiments are 1125 and 2880.
 
After LBPTOP feature extraction, Support Vector Machine (SVM) with Radial Basis Function (RBF) is utilized for training classifiers with extracted feature vectors. The RBF kernel is employed so that the classifier SVM \change[CL]{could find optimized}{optimizes} decision hyper-planes for these multi-class recognition problems in the infinite similarity spaces of RBF kernel rather than original small feature spaces \change[JS]{e.g.}{i.e.} 1125-D feature of CASME II samples and 2880-D features of SMIC samples. So far, the cost parameter (c) and Gaussian Kernel Bandwidth (g) are set to $10^4$ and $0.5$ respectively, as shown in Table \ref{sec-eds:tab4}. 
\add[CL]{The kernel bandwidth is kept at a default value $0.5$ and the cost parameter (c) is set to a very large value to prevent misclassification of inter-class samples from the same subject.}

As facial expressions and identities are inseparable characteristics of all video samples, the special Leave-One-Subject-Out (LOSO) protocol \cite{lucey2010extended}\cite{littlewort2006dynamics}\cite{li2013spontaneous}\cite{pfister2011recognising} is \add[CL]{also} enforced \remove[CL]{in the machine learning stage to ensure that recognition of facial expressions is independent of identities}{to prevent subject identities interfere with classification of subtle emotions}. LOSO is an exhaustive cross-validation technique, \remove[CL]{which partitions}{partitioning corpus} testing and training \change[CL]{corpus}{sets} according to subject identities. It requires samples of the testing corpus from one subject and those of the training corpus from the other subjects. For example, if Subject 1 of CASME II corpus is chosen as the test subject, all nine video samples of the subject are preserved for testing. The other video samples (238) are used for training the classifier SVM-RBF. These processes are folded until every subject is employed in the test corpus \remove[CL]{at least} once. There are 26 subjects in the CASME II database hence 26 folds of cross-validation in the LOSO protocol are carried out. All 26 combinations of training and testing samples can be inferred from Table 2 in \cite{lengo2014imbalance}. For \change[CL]{evaluating the performances of systems}{performance metric}, \remove[CL]{the} Accuracy (ACC), F1-score (F1), Recall (RR) and Precision (PR) Rates are \change[CL]{used in the evaluation process of}{employed in} this work due to class imbalance or high skewness \cite{lengo2014imbalance} in \remove[CL]{subtle emotion databases like} CASME II and SMIC . With respect to each class "c", the $\text{F1}_{\text{c}}$, $\text{R}_{\text{c}}$ and $\text{P}_{\text{c}}$ are computed from the number of true positive ($\text{TP}_{\textbf{c}}$), false negative ($\text{FN}_\text{c}$) and false positive ($\text{FP}_\textbf{c}$) as follows.
\[
	\text{RR}_{\text{c}} = \frac{\text{TP}_{\textbf{c}}}{\text{TP}_{\textbf{c}}+\text{FN}_\textbf{c}} \quad 
	\text{PR}_{\text{c}} = \frac{\text{TP}_{\textbf{c}}}{\text{TP}_{\textbf{c}}+\text{FP}_\textbf{c}} \quad
	\text{F1}_{\text{c}} = \frac{2\text{R}_\text{c}\text{P}_\text{c}}{\text{R}_{\text{c}}+\text{P}_{\text{c}}}
\]
For overall performance on the whole corpus, F1-score, Recall and Precision rates of the dataset are taken as the mean of all respective values of its classes; meanwhile, the overall accuracy (ACC) is computed as $\frac{\sum_{c} \text{TP}_\textbf{c} }{\sum_{c} \text{TP}_\textbf{c} + \text{FN}_\textbf{c}}$. Table \ref{sec-eds:tab2} shows average evaluation results across all folds and all classes; meanwhile, Table \ref{sec-eds:tab1} displays the average evaluation results across all folds for each class. Note that the CASME II paper \cite{yan2014casme} employs a different evaluation protocol Leave-one-video-out (LOVO); therefore, Yan et al. \cite{yan2014casme} reports much different and higher accuracy rates for 5-class emotion recognition with the CASME II database in Table 4 of \cite{yan2014casme}. These results of LOVO are incomparable to those of LOSO in Table \ref{sec-eds:tab1}. As LOVO involves samples belonging to the same subject in both training and testing corpora, it may bias the recognition results. 

\begin{figure*}[t]
	\begin{subfigure}{\columnwidth}
		\IfFileExists{experimentalResultOnCASME2_f1Score_plot.tex}
		{\input{experimentalResultOnCASME2_f1Score_plot.tex}}
		{\includegraphics{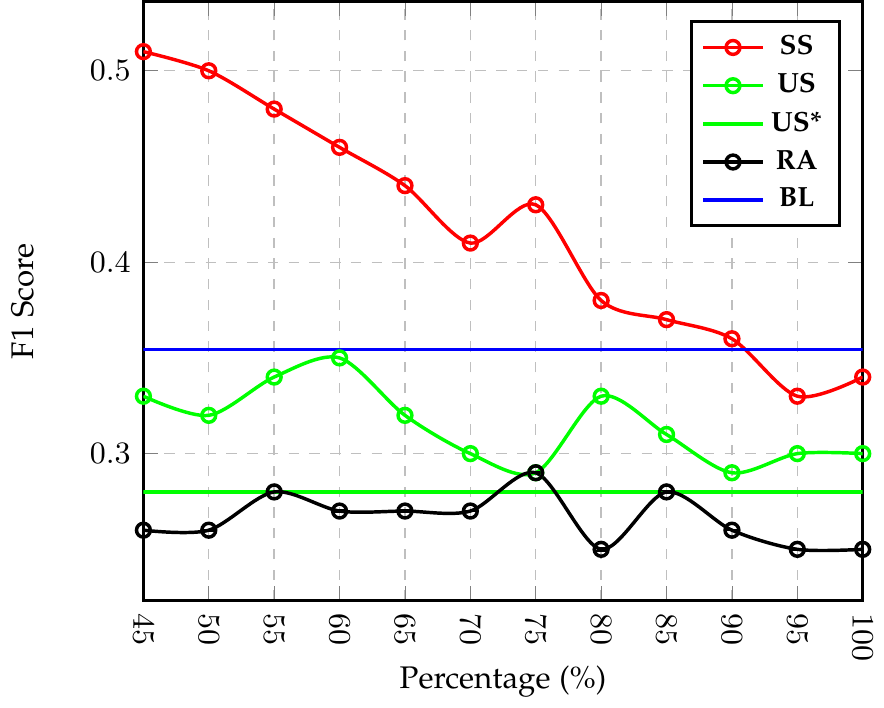}}
		\caption{F1 Scores of different approaches on CASME II}
		\label{sec-eds:fig1}
	\end{subfigure}
	\begin{subfigure}{\columnwidth}
		\IfFileExists{experimentalResultOnSMIC_f1Score_plot.tex}
		{\input{experimentalResultOnSMIC_f1Score_plot.tex}}
		{\includegraphics{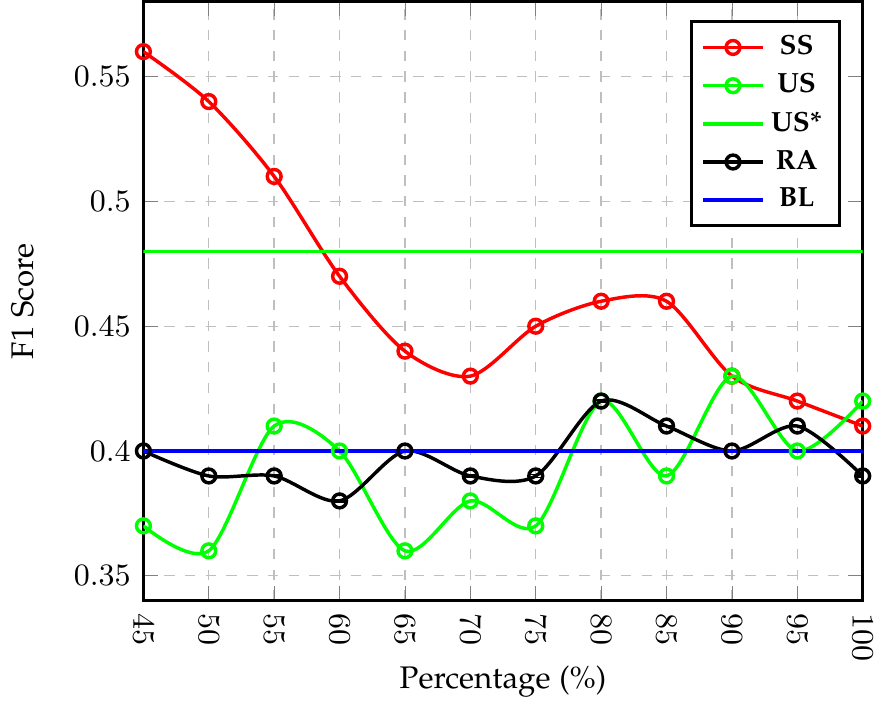}}
		\captionof{figure}{F1 Scores of different approaches on SMIC}
		\label{sec-eds:fig4}
	\end{subfigure}	
	\\
	\begin{subfigure}{\columnwidth}
		\IfFileExists{experimentalResultOnCASME2_recall_plot.tex}
		{\input{experimentalResultOnCASME2_recall_plot.tex}}
		{\includegraphics{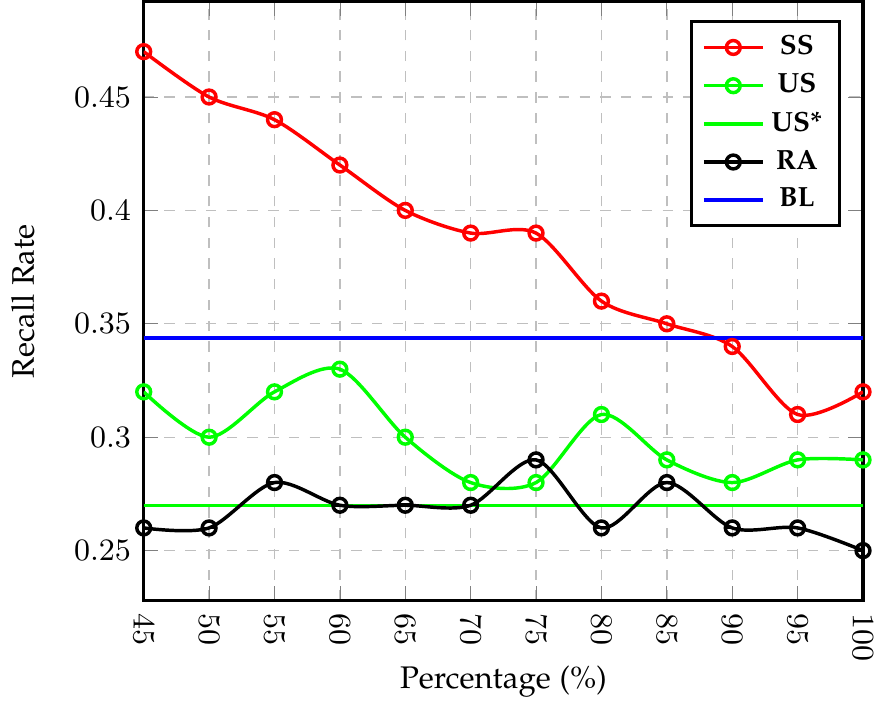}}
		\caption{Recall Rates of different approaches on CASME II}
		\label{sec-eds:fig2}
	\end{subfigure}	
	\begin{subfigure}{\columnwidth}
		\IfFileExists{experimentalResultOnSMIC_recall_plot.tex}
		{\input{experimentalResultOnSMIC_recall_plot.tex}}
		{\includegraphics{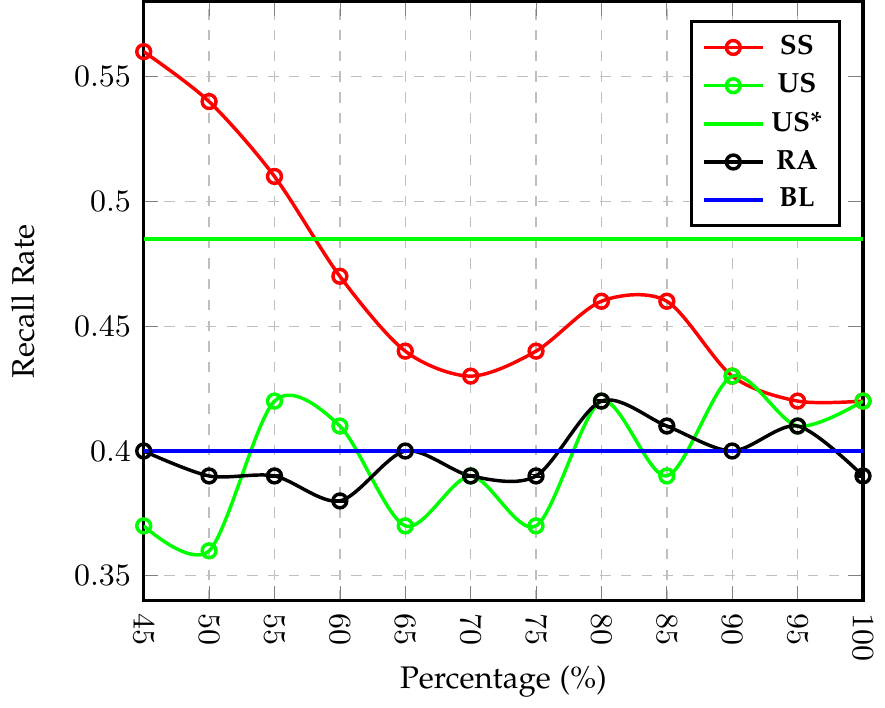}}
		\captionof{figure}{Recall Rates of different approaches on SMIC}
		\label{sec-eds:fig5}
	\end{subfigure}
	\\
	\begin{subfigure}{\columnwidth}
		\IfFileExists{experimentalResultOnCASME2_precision_plot.tex}
		{\input{experimentalResultOnCASME2_precision_plot.tex}}
		{\includegraphics{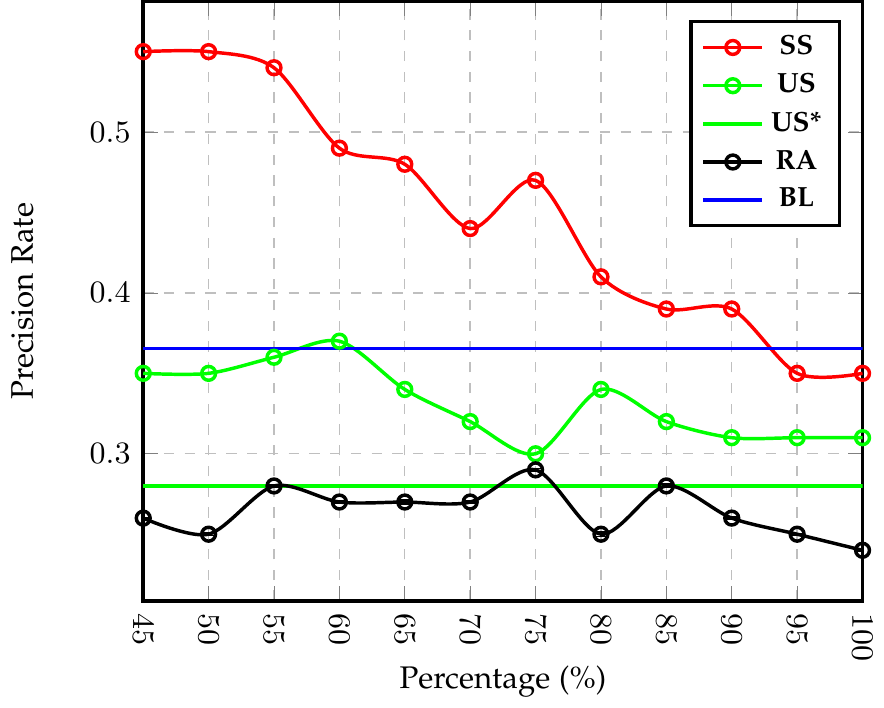}}
		\caption{Precision Rates of different approaches on CASME II}
		\label{sec-eds:fig3}
	\end{subfigure}
	\begin{subfigure}{\columnwidth}
		\IfFileExists{experimentalResultOnSMIC_precision_plot.tex}	
		{\input{experimentalResultOnSMIC_precision_plot.tex}}
		{\includegraphics{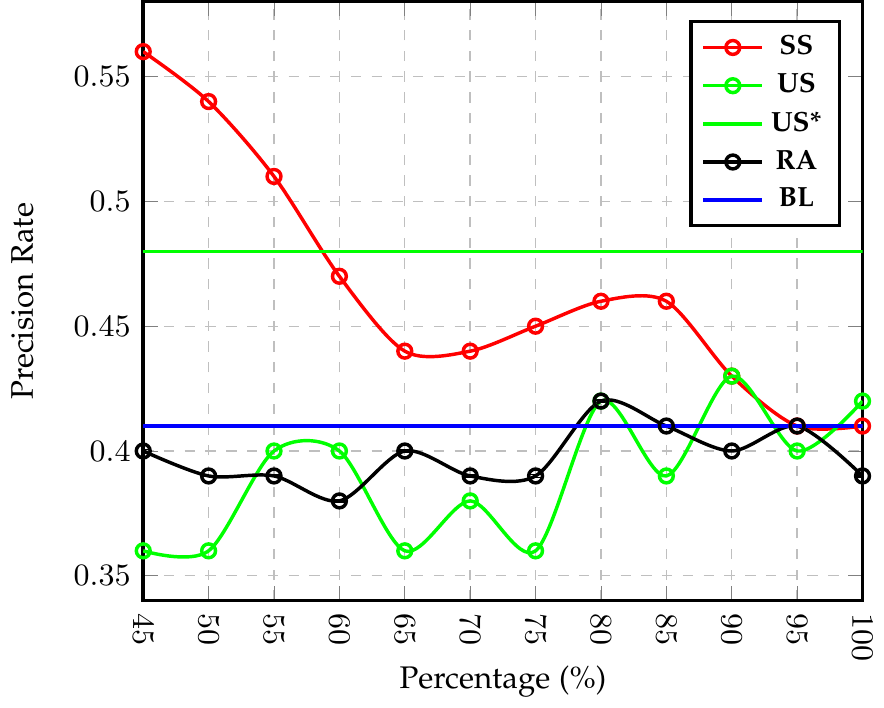}}
		\captionof{figure}{Precision Rates of different approaches on SMIC}
		\label{sec-eds:fig6}
	\end{subfigure}	
	\caption{Performance metrics (F1 score, Recall and Precision Rates) of various temporally sampling methods with respect to a range of percentages of preserved frames (45\%-100\% original frame-length) on CASME II and SMIC databases}
\end{figure*}

\subsection{Sparse Sampling vs Uniform Sampling}
\label{sec-eds:subsec-svu}

In Table \ref{sec-eds:tab1}, average performances across all LOSO folds and classes of five \textbf{SS}, \textbf{US}, \textbf{US*}, \textbf{RA}, and \textbf{BL} approaches are compared. Furthermore, Figures \ref{sec-eds:fig1}, \ref{sec-eds:fig2}, and \ref{sec-eds:fig3} show plots of F1-score, Recall Rate and Precision Rate of 5-class subtle emotion recognition on CASME II along percentages (p) of preserved frames (45\% - 100\%) so do Figures \ref{sec-eds:fig4}, \ref{sec-eds:fig5} and \ref{sec-eds:fig6} for the 3-class recognition on SMIC. These plots allow visualization of performances of the approaches with respect to different amounts of redundancy. 

Among all evaluation results shown in Figures \ref{sec-eds:fig1}-\ref{sec-eds:fig6}, the best performances of the recognition system occur at the minimum redundancy $p=45\%$ for both CASME II 5-class and SMIC 3-class recognition. Corresponding F1 scores, Precision and Recall rates are listed at the "\textbf{SS}" row for both CASME II and SMIC databases. It is experimentally validated that the proposed \textbf{SS} approach effectively removes dynamic redundancies compared to any other methods. In accordance with Table \ref{sec-eds:tab1} and Figures \ref{sec-eds:fig1}-\ref{sec-eds:fig6}, the \textbf{SS} approach improves F1, PR and RR by 40\% and ACC by 30\% over the second-best performances i.e. \textbf{BL} or \textbf{US} for CASME II corpus. Meanwhile, F1, RR, PR and ACC of the \textbf{SS} approach are improved by 20\%-25\% over the second best result (of the \textbf{US*} approach) for the SMIC corpus. 
The performance gap between \textbf{SS} and \textbf{US} approaches confirms the existence of sparse and redundant dynamics in video samples of micro-expressions. Moreover, the proposed \textbf{SS} approach effectively preserves sparse dynamics i.e. emotional expressions, and removes redundant ones i.e. neutral expressions. When redundancies are removed from dynamics of expressions, it increases inter-class and reduces intra-class distances between samples of different emotional classes. As each class consists of video samples from most if not all participating subjects, its intra-class distances are large and inter-class distances are small due to redundant facial identities or neutral expressions. Through elimination of these redundancies, discrimination of video samples, reconstructed by the \textbf{SS} approach, w.r.t different emotions is better than that of original samples. Therefore, more compact but meaningful sequences in turn lead to better recognition performances, as shown in Table \ref{sec-eds:tab1}.

The performance of the \textbf{US} approach is worse than the \textbf{SS} and equivalent to \textbf{BL} as shown in Table \ref{sec-eds:tab1} since TIM arbitrarily removes both discriminative and redundant features by regularly sampling dynamics along a Laplacian manifold. Moreover, the \textbf{US} only has better performances than \textbf{RA} by a small margin 0.02-0.03 in terms of F1, RR, PR and ACC. The \textbf{US} and \textbf{US*} rows show experimental results of uniform sampling approach with different frame-lengths of generated sequences. Results of these experiments, shown in Figures \ref{sec-eds:fig1} and \ref{sec-eds:fig6}, show that the best performance of \textbf{US} (green) lines occur at $p=60\%$ for CASME II and at $p=90\%$ for SMIC. The best F1, PR and RR of the \textbf{US} approach are presented at the second row of Table \ref{sec-eds:tab1}. 
In Yan et al. \cite{yan2014casme} and Pfister et al. \cite{pfister2011recognising}, TIM is used for opposite purposes. While CASME II's sequences are lengthened or extrapolated to 150 frames for every sequence in \cite{yan2014casme}, SMIC sequences are shortened or interpolated to 10 frames in \cite{pfister2011recognising}. The \textbf{US*} results in Table \ref{sec-eds:tab1} show recognition performances with respect to those uniform sampling parameters. However, regardless of extrapolation or interpolation, both the \textbf{US} and \textbf{US*} approaches are inferior to the proposed \textbf{SS} approach in all performance measures, as shown in Table \ref{sec-eds:tab1} and Figures \ref{sec-eds:fig1}-\ref{sec-eds:fig6}. Overall, the proposed sparse sampling approach greatly boosts performances of subtle emotion recognition tasks. \remove[CL]{which are evaluated on all publicly available databases of spontaneous micro expressions like CASME II and SMIC.}

\subsection{Sparse Sampling vs Other Methods}
\label{sec-eds:subsec-som}
Yan et al. \cite{yan2014casme} uses the leave-one-video-out (LOVO) protocol for evaluating performances of 5-class subtle emotion recognition on CASME II database instead of the common LOSO protocol. Though the merit of LOVO evaluation is questionable due to involvement of same subjects in both training and testing phases, we additionally carry out the evaluation based on LOVO for directly comparing the proposed (\textbf{SS}) approach with Yan el al. \cite{yan2014casme} (\textbf{YA}). The comparison is valid as all experimental parameters of the proposed approach \textbf{SS} for CASME II in Table \ref{sec-eds:tab5} are also used in \cite{yan2014casme}. It is noted that the \textbf{SS} approach in the table only uses 45\% of original frame-lengths. Table \ref{sec-eds:tab5} shows a significant average improvement in ACC, F1 score, Precision and Recall rates of \textbf{SS} over \textbf{YA} with the LOVO evaluation protocol. Furthermore, the improvement is also observed in four over five individual classes 'Disgust', 'Happiness', 'Surprise' and 'Repression'. In the 'Others' class, F1 scores of both approaches are equivalent i.e. no improvement of \textbf{SS} over \textbf{YA}. It is due to skewed sample distribution towards the 'Others' class in CASME II database, shown in Table \ref{sec-eds:tab3} of \cite{lengo2014imbalance}. Hence, hyperplanes of SVM classifiers are over-fitted with respect to a cluster of the 'Others' samples. As the biased classifier tends to detect large numbers of true positives as well as false positive samples of \add[CL]{the} 'Others' class, its recall rate in \textbf{YA} is high but precision rate is low. Meanwhile in the \textbf{SS} approach, the recall rate drops by 0.03 and precision rate increases by 0.04 from corresponding values of \textbf{YA}. \textbf{SS} partially removes the bias toward the 'Others' class and improves F1 scores, R and P rates of the other categories.

\begin{table}[t]
	\scriptsize
	\centering	
	\caption{Leave-one-video-out evaluation on CASME II database of Sparse Sampling (\textbf{SS}) approach and Yan et al. \cite{yan2014casme} (\textbf{YA})}
	\label{sec-eds:tab5}
	\resizebox{\columnwidth}{!}{
	\begin{tabular}{ccccccccccc}
		\toprule
		& \multicolumn{4}{c}{Average} & \multicolumn{3}{c}{Others (O)} & \multicolumn{3}{c}{Disgust (D)} \\
		\cmidrule(r){2-5} \cmidrule(r){6-8} \cmidrule(r){9-11}
		& ACC & F1 & RR & PR & F1 & RR & PR & F1 & RR & PR \\
		\midrule 
		\textbf{SS} & .72 & .71 & .70 & .72 & .75 & .76 & .74 & .74 & .72 & .75 \\
		\textbf{YA} & .64 & .59 & .58 & .60 & .75 & .79 & .70 & .60 & .55 & .66 \\ 
		\textbf{(SS-YA)/SS} & +12\% & +20\% & +21\% & +20\% & 0\% & -4\% & +4\% & +23\% & +30\% & +14\% \\ \midrule
		&  & \multicolumn{3}{c}{Happiness (H)} & \multicolumn{3}{c}{Surprise (S)} & \multicolumn{3}{c}{Repression (R)} \\
		\cmidrule(r){3-5} \cmidrule(r){6-8} \cmidrule(r){9-11}
		&  & F1 & RR & PR & F1 & RR & PR & F1 & RR & PR \\ \midrule
		\textbf{SS} &  & .56 & .61 & .51 & .82 & .80 & .83 & .67 & .59 & .76 \\		
		\textbf{YA} &  & .47 & .45 & .48 & .62 & .64 & .59 & .51 & .48 & .54 \\
		\textbf{(SS-YA)/SS} &  & +19\% & +33\% & +6\% & +33\% & +25\% & +41\% & +31\% & +23\% & +41\% \\ \bottomrule
	\end{tabular}}
\end{table}

Besides bench-marking against the baseline method of the CASME II database \cite{yan2014casme}, we also compare the proposed approach to other more recent methods from Huang et al. \cite{huang2015facial}, Oh et al. \cite{oh2015monogenic}, Liong et al. \cite{liong2014subtle}, Wang et al. \cite{wang2015efficient} and Le Ngo et al. \cite{lengo2014imbalance} with respect to the \change[CL]{as they employ similar}{above} evaluation protocols. Table \ref{sec-eds:tab6} compares performances of the proposed "Sparse Sampling" method with other mentioned state-of-the-art methods. Methods of \cite{oh2015monogenic},\cite{liong2014subtle} and \cite{lengo2014imbalance} are re-implemented; hence, their results in Table \ref{sec-eds:tab6} are re-evaluated according to parameters of the proposed method with the LOSO protocol. Meanwhile, Huang et al. \cite{huang2015facial} provides confusion tables for both recognition evaluation on CASME II and SMIC; therefore, F1-score, recall and precision rates are computed directly from the tables without re-implementation. In general, the proposed approach outperforms most of state-of-the-art methods for both CASME II and SMIC databases except Huang et al. \cite{huang2015facial} on the CASME II database. Furthermore, "Sparse Sampling" is the only available method 
that aims to select significant temporal dynamics before feature extraction; meanwhile, the other methods solely focus on designing better features and classifiers. \change[CL]{It}{Experimental results} shows that proper selection of facial dynamics\remove[CL]{can} significantly improves recognition rates of subtle expressions while reducing computational efforts.  

\begin{table}[h]
	\caption{Performance comparison in 5-class recognition for CASME II corpus and 3-class recognition for SMIC against other state-of-the-art methods with the LOSO protocol}	
   	\resizebox{\columnwidth}{!}{
	\begin{tabular}{l*{8}{c}}
		\toprule
		& \multicolumn{4}{c}{\textbf{CASME II}} & \multicolumn{4}{c}{\textbf{SMIC}} \\
		\cmidrule(r){2-5} \cmidrule(r){6-9}
		& ACC & F1 & R & P & ACC & F1 & R & P \\
		\midrule
		\textbf{Sparse Sampling} & \textbf{.49} & \textbf{.51} & \textbf{.47} & \textbf{.55} & \textbf{.58} & \textbf{.60} & \textbf{.60} & \textbf{.60} \\
		Huang et at. \cite{huang2015facial} & \textit{.59} & \textit{.57} & \textit{.51} & \textit{.65} & \textit{.57} & \textit{.58} & \textit{.58} & \textit{.59} \\		
		Oh et al.\cite{oh2015monogenic} & .46 & .43 & .35 & .55 & .34 & .35 & .35 & .34 \\
		Liong et al.\cite{liong2014subtle} & .42 & .38 & .36 & .41 & .53 & .54 & .55 & .53 \\
		Wang et al. \cite{wang2015efficient} & .46 & .38 & .32 & .47 & .38 & .39 & .40 & .38 \\
		Le et al.\cite{lengo2014imbalance} & .44 & .33 & .53 & .29 & .44 & .47 & .74 & .40 \\		
		Yan et al.\cite{yan2014casme} & .38 & .35 & .34 & .36 & {\scriptsize N/A} & {\scriptsize N/A} & {\scriptsize N/A} & {\scriptsize N/A} \\
        Pfister et al. \cite{pfister2011recognising} & {\scriptsize N/A} & {\scriptsize N/A} & {\scriptsize N/A} & {\scriptsize N/A} & .40 & .40 & .40 & .41 \\
		\bottomrule
	\end{tabular}
	\label{sec-eds:tab6}
    }
\end{table}



\section{Conclusion}
\label{sec-con}
This work is the first ever endeavour to analyz\change[CL]{ing}{e} the dynamics of spontaneous subtle emotions and to learn \add[CL]{their} sparse structures. \remove[CL]{of the dynamics} Knowledge of these subtle dynamics and the sparsity of data enable spontaneous micro-expressions to be described more discriminately through pre-processing video samples. Both sparse (DMDSP) and uniform sampling (TIM) are compared in theory as principles for removing dynamic redundancy 
and then both are analyzed in the temporal and spectral domains. The \change[CL]{theories and}{experimental} analyses show that the sparse sampling approach is more accurate and consistent than the uniform counterpart. Moreover, when 
compared against the other state-of-the-art methods in recognizing spontaneous subtle emotion, the performances of the proposed method are very competitive against the state-of-the-art e.g. the best for the SMIC database and the second for CASME II. Therefore, these experimental results confirm the existence of redundancies in the dynamics of spontaneous subtle emotions, and their removals by sparse sampling cause micro-expressions to be more distinctive and recognizable.

\section{Acknowledgements}

The authors (ACLN, JS, RP) thank S\'{e}bastien Marcel for suggesting to use the F1 score instead of accuracy as a measure, Norman Poh for sharing about his biometrics work involving DMD, and Jonathon Chambers for discussions on DMD and DMDSP.
The research and collaboration discussions with S\'{e}bastien, Norman and Jonathon were funded by TM (Telekom Malaysia) under the projects UbeAware (MMUE/130152) and 2beAware (MMUE/140098).
RP gratefully acknowledges the support by the UK Engineering \& Physical Sciences Research Council (EPSRC) under the project Signal Processing Solutions for the Networked Battlespace (EP/K014307/1$\sim$2) and the MoD University Defence Research Collaboration (UDRC) in Signal Processing.


\bibliographystyle{IEEEtran}
\bibliography{IEEEabrv,ref}

\begin{IEEEbiography}[{
\includegraphics[width=1in,height=1.25in,clip,keepaspectratio]{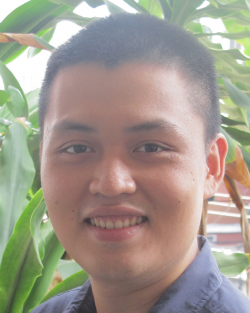}}]{Anh-Cat Le-Ngo, Ph.D.} was born in Ho Chi Minh city, Vietnam. He has finished his doctorate training in Computer Vision and Image Processing in University of Nottingham in 2015. Since 2013, he has been working as a postdoc in the UbeAware and 2beAware projects under the guidance of Prof. Raphael C-.W. Phan. His main research interests are computer vision systems (i.e. micro-expression recognition system (MERS), advance driving assistance system (ADAS), etc), signal \& image processing mathematics (i.e. monogenic signal, analytic signal, Hilber \& Riesz transforms, Dynamic Mode Decomposition, etc) and machine learning (i.e. Long Short Term Memory (LSTM), Recurrent Neural Network (RNN), Selective Transfer Machine (STM), etc).
\end{IEEEbiography}

\begin{IEEEbiography}[{
\includegraphics[width=1in,height=1.25in,clip,keepaspectratio]{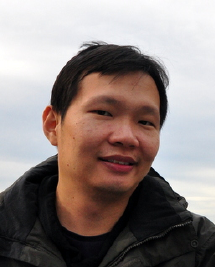}
}]{John See, Ph.D.} received his PhD in Computer Science, MEngSc and BEng degrees from Multimedia University (MMU), Malaysia. He is currently working as a Senior Lecturer and leader of the Pattern Recognition sub-cluster of the Centre for Visual Computing at Multimedia University, Malaysia. His research interests covers a diverse range of topics in computer vision and pattern recognition, particularly in the domain of video-based biometrics, visual surveillance, affective computing, image aesthetics and deep learning. His passion remains in designing effective and efficient algorithms for visual recognition tasks.
\end{IEEEbiography}

\begin{IEEEbiography}[{
\includegraphics[width=1in,height=1.25in,clip,keepaspectratio]{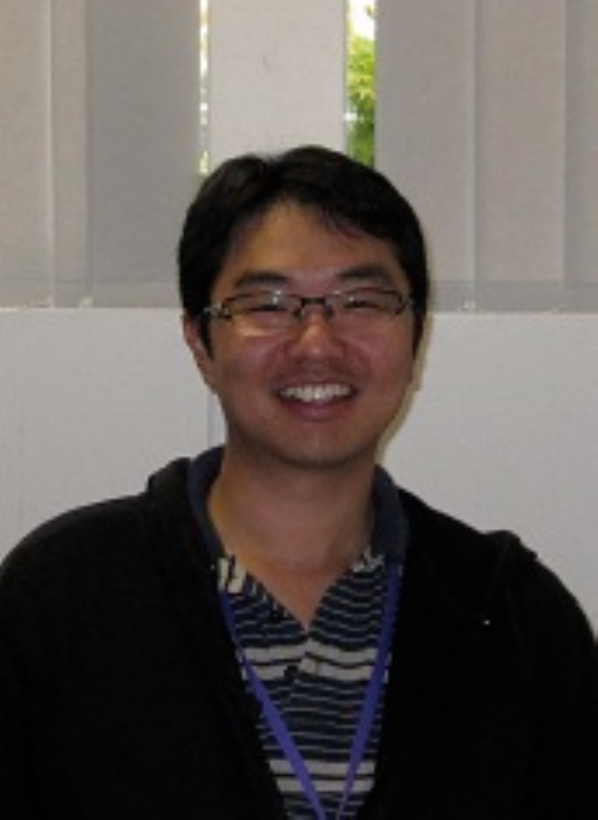}
}]{Raphael C.-W. Phan, Ph.D.} holds the full chair in security engineering at the Faculty of Engineering, Multimedia University (MMU).  Prior to joining MMU, he worked at British, Swiss and Australian universities. He has led projects funded by the UK Engineering \& Physical Sciences Research Council (EPSRC), UK Ministry of Defence (MoD), as well as the Malaysian government and industry.  Raphael's research passion spans the breadth of security \& privacy, from cryptography and security protocols to subtle and/or hidden emotions: essentially aiming to see beyond the unseen.  He was co-designer of the BLAKE hash function, one of the five finalists of the US National Institute of Standards \& Technology (NIST) SHA-3 Hash Function Competition.
Raphael is regularly invited to serve in the technical program committees of peer-reviewed security conferences, and is  Program Chair of Mycrypt 2016, the first such conference of its kind with focus on malicious and/or out-of-the-box cryptographic paradigms.
\end{IEEEbiography}

\end{document}